\documentclass[letterpaper]{article} 
\usepackage{aaai2026}  
\usepackage{times}  
\usepackage{helvet}  
\usepackage{courier}  
\usepackage[hyphens]{url}  
\usepackage{graphicx} 
\usepackage{natbib}  
\usepackage{caption} 
\usepackage{algorithm}
\usepackage{algorithmic}

\usepackage{newfloat}
\usepackage{listings}
\DeclareCaptionStyle{ruled}{labelfont=normalfont,labelsep=colon,strut=off} 
\floatstyle{ruled}
\newfloat{listing}{tb}{lst}{}
\floatname{listing}{Listing}
\title{
DISCODE: Distribution-Aware Score Decoder for\\
Robust Automatic Evaluation of Image Captioning
}
\author{
Nakamasa Inoue\textsuperscript{\rm 1}\equalcontrib, Kanoko Goto\textsuperscript{\rm 1}\equalcontrib, Masanari Oi\textsuperscript{\rm 1}, Martyna Gruszka\textsuperscript{\rm 1},\\Mahiro Ukai\textsuperscript{\rm 1}, Takumi Hirose\textsuperscript{\rm 1}, Yusuke Sekikawa\textsuperscript{\rm 2}
}
\affiliations{
\textsuperscript{\rm 1}Institute of Science Tokyo\quad \\
\textsuperscript{\rm 2}Denso IT Laboratory
}
\usepackage{bibentry}
\newcommand{\red}[1]{{\color{red}#1}}

\usepackage[cache=false]{minted}
\usepackage{tcolorbox}
\usepackage{colortbl}
\usepackage{xcolor}
\usepackage{amsmath}
\usepackage{algorithmic}
\usepackage{algorithm}
\usepackage{bm}
\usepackage{amssymb} 
\usepackage{booktabs}
\usepackage{comment}
\usepackage{multirow}
\usepackage{framed}
\def\red#1{{\color{red} #1}}

\usepackage{url}
\makeatletter
\newcommand{\figcaption}[1]{\def\@captype{figure}\caption{#1}}
\newcommand{\tblcaption}[1]{\def\@captype{table}\caption{#1}}
\makeatother

\newcommand{\argmin}{\mathop{\rm argmin}\limits}

\newcount\K
\def\bla#1{
\K=0 \loop\ifnum\K<#1
{\textcolor[gray]{0.9}{{\it bla bla bla bla bla bla bla bla bla bla bla bla bla bla bla}}}
\advance\K by1\repeat
}

\def\paragraph#1{\noindent \textbf{#1.}}
\usepackage{colortbl,array,xcolor}

\def\sref#1#2{\hyperref[#1]{\ref*{#1}#2}}
\usepackage{amsthm}
\usepackage{cuted}
\begin{document}
\maketitle
\def\ATT{\text{\scalebox{0.8}{ATT}}}
\def\LVLM{\text{\scalebox{0.8}{\hspace{1pt}LVLM}}}

\begin{abstract}
Large vision-language models (LVLMs) have shown impressive performance across a broad range of multimodal tasks.
However, robust image caption evaluation using LVLMs remains challenging, particularly under domain-shift scenarios.
To address this issue, we introduce the {\textbf{Distribution-Aware Score Decoder (DISCODE)}}, a novel finetuning-free method that generates robust evaluation scores better aligned with human judgments across diverse domains.
The core idea behind DISCODE lies in its test-time adaptive evaluation approach,
which introduces the Adaptive Test-Time (ATT) loss, leveraging a Gaussian prior distribution to improve robustness in evaluation score estimation.
This loss is efficiently minimized at test time using an analytical solution that we derive.
Furthermore, we introduce the \textbf{Multi-domain Caption Evaluation (MCEval) benchmark}, a new image captioning evaluation benchmark covering six distinct domains, designed to assess the robustness of evaluation metrics.
In our experiments, we demonstrate that DISCODE achieves state-of-the-art performance as a reference-free evaluation metric across MCEval and four representative existing benchmarks.
\end{abstract}

\section{Introduction}
\label{sec:intro}
Developing automatic evaluation metrics that closely correlate with human judgments is essential for advancing toward more human-centric artificial intelligence.
For image captioning tasks, significant efforts have been devoted to designing automatic evaluation metrics, beginning with traditional methods such as BLEU~\cite{Papineni2002BLEU} and CIDEr~\cite{Vedantam2015CIDEr}.
Nevertheless, accurate evaluation remains challenging due to the inherent variability and subjectivity of natural language descriptions.

\begin{figure}
\centering
\includegraphics[width=\linewidth]{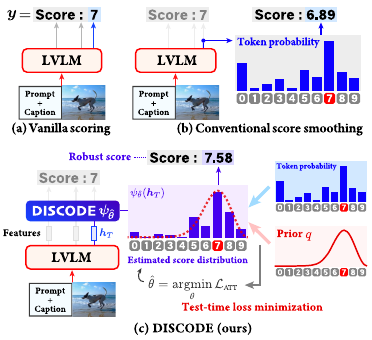}
\caption{
Scoring with DISCODE.
(a) Vanilla scoring: LVLM generates raw scores.
(b) Score smoothing: Expected score is computed from the token probability distribution.
(c) DISCODE (ours): Score distribution is robustly estimated from the decoder feature $\bm{h}_{T}$ by minimizing the ATT loss $\mathcal{L}_{\ATT}$, which leverages a Gaussian prior $q$ at test time.}
\label{fig:top}
\end{figure}

Recently, large vision-language models (LVLMs) have demonstrated substantial improvements in image-text alignment tasks~\cite{liu2023llava, Wang2024Qwen2VL, Chen2024InternVL, lu2024deepseekvl}. 
To perform accurate numerical evaluation using LVLMs, 
state-of-the-art methods such as FLEUR~\cite{Lee2024FLEUR} and G-VEval~\cite{Tong2025GVEval} leverage the \textit{score smoothing} technique, which generates real-valued scores by estimating the score distribution based on the output token probability distribution assigned by LVLMs.
However, robustly aligning generated scores with human judgments remains difficult, especially in domain-diverse scenarios.

We hypothesize that this difficulty stems from the discrepancy between the token probability distribution and the human evaluation score distribution, particularly in terms of unimodality.
Due to the central limit theorem, human evaluation scores naturally tend to follow a Gaussian distribution.
In contrast, token probability distributions typically do not exhibit Gaussian behavior and instead show certain biases, such as symbolic bias, where specific tokens are disproportionately frequent (as illustrated in Figure~\ref{fig:observation}). 
This discrepancy becomes more pronounced in certain visual domains such as paintings and abstract sketches because these domains often involve subjective interpretations and greater semantic ambiguity.
This motivates us to propose a novel method employing a Gaussian prior distribution to enhance the robustness of evaluation scores, along with a new benchmark dataset encompassing diverse visual domains.

Specifically, this paper makes two major contributions. First, we propose the Distribution-Aware Score Decoder (\textbf{DISCODE}), a novel test-time adaptive decoder for LVLMs that generates robust scores by leveraging a Gaussian prior distribution.
Second, we introduce the Multi-domain Caption Evaluation (\textbf{MCEval}) benchmark, a new image captioning evaluation benchmark spanning six visual domains.
In our experiments, we demonstrate that DISCODE achieves state-of-the-art performance on MCEval as well as four representative benchmarks: Flickr8k-Expert~\cite{Hodosh2013Flickr8kEXCF}, Flickr8k-CF~\cite{Hodosh2013Flickr8kEXCF}, Composite~\cite{Aditya2015Composite}, and Pascal-50S~\cite{Vedantam2015CIDEr}. Our contributions are summarized as follows.
\begin{figure}
\centering
\includegraphics[width=\linewidth]{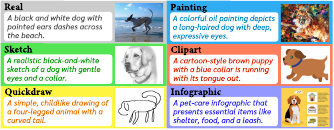}
\caption{MCEval benchmark. We provide a human evaluation dataset covering six visual domains for assessing the robustness of evaluation metrics.}
\label{fig:topb}
\end{figure}

\noindent \textbf{1)\hspace{2pt}Technical Contribution.}
We propose DISCODE, a novel decoder for LVLM-based image captioning evaluation.
As shown in Figure~\ref{fig:top},
DISCODE minimizes the Adaptive Test-Time (ATT) loss, which measures discrepancy from a Gaussian prior distribution, enabling LVLMs to function as a robust evaluation metric.
Furthermore, we derive a closed-form analytical solution to the loss minimization problem, leading to an efficient implementation.

\noindent \textbf{2)\hspace{2pt}Dataset Contribution.} We introduce MCEval, a new dataset for benchmarking the robustness and generalizability of evaluation metrics.
Our dataset consists of 18,000 image-text pairs spanning six visual domains, as shown in Figure~\ref{fig:topb}, along with human evaluation ground-truth labels.

\section{Related Work}
\label{sec:related_work}

Image caption evaluation metrics can be divided into two categories: reference-based and reference-free metrics.

\noindent \textbf{Reference‑based metrics.} This approach quantifies caption quality by comparing candidate captions with human-written references.
Classical metrics such as BLEU~\cite{Papineni2002BLEU}, ROUGE~\cite{Lin2004ROUGE}, and METEOR~\cite{Banerjee2005METEOR} rely on $n$‑gram overlap.
CIDEr~\cite{Vedantam2015CIDEr} incorporates TF–IDF weighting to emphasize consensus, and SPICE~\cite{Anderson2016SPICE} captures semantic structure by matching scene graphs. 
With advances in pretrained language models, caption similarity has become measurable in representational space rather than surface form, leading to embedding‑based metrics such as BERTScore~\cite{Zhan2020BERTScore, Yi2020BERTScorepp}, BARTScore~\cite{Weizhe2021BARTScore}, TIGEr~\cite{Jiang2019TIGEr} and ViLBERTScore~\cite{Lee2020ViLBERTScore}.
Finetuning-based metrics have further enhanced alignment with human judgments; examples include FAIEr~\cite{wang2021faier}, Polos~\cite{Wada2024PolosPolaris}, CAMScore ~\cite{cui2025evaluating}, DENEB~\cite{Matsuda2024DENEB}and MLF~\cite{gomes-etal-2025-evaluation}.

\noindent \textbf{Reference‑free metrics.} To evaluate image captions without relying on human-written references, reference-free metrics leverage pre-trained vision-language models.
CLIP-Score~\cite{Hessel2021CLIPS} is a representative metric that measures the alignment between images and texts using CLIP embeddings~\cite{radford2021CLIP}.
PAC-S~\cite{Sarto2023PACS, Sarto2024PACScorepp} introduced positive-augmented contrastive learning, leveraging image generators to more precisely align image and text embeddings.
HiFi-Score~\cite{yao2024hifiscore} introduced graph-based matching.
Recent studies have demonstrated that LVLMs can serve as effective evaluators.
To generate accurate real-valued scores,
FLEUR~\cite{Lee2024FLEUR} introduced improved score smoothing for LLaVA-1.5~\cite{liu2023llava}.
G-VEval~\cite{Tong2025GVEval} designed chain-of-thought prompts~\cite{wei2022chainofthought} for caption evaluation using GPT-4o.
These sophisticated approaches rely on the score smoothing technique~\cite{Liu2023GEval}, which estimates the score distribution based on the output token probability distribution. However, token probability distributions often exhibit unintended biases, such as symbolic bias, causing them to become non-unimodal.
DISCODE enhances evaluation robustness by addressing this issue through minimization of the ATT loss, a novel test-time loss formulated with a Gaussian prior distribution.

\begin{figure*}[ht]
\centering
\includegraphics[width=0.9\linewidth]{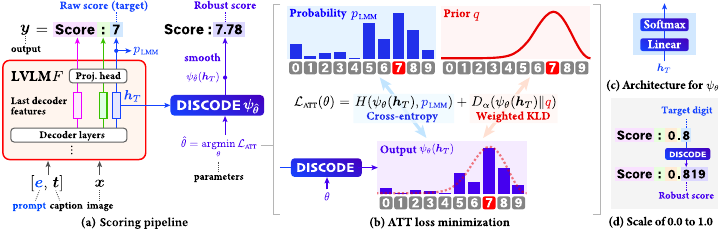}
\caption{Overview of DISCODE.
(a) Scoring pipeline where DISCODE $\psi_{\hat{\theta}}$ is applied to the decoder feature $\bm{h}_{T}$ to generate robust scores.
(b) ATT loss to minimize cross entropy and weighted Kullback-Leibler divergence (KLD).
(c) Architecture for $\psi_{\theta}$, which consists of a linear layer and a softmax function.
(d) Target digit for evaluation on a scale of 0.0 to 1.0.
}
\label{fig:discode}
\end{figure*}

\section{Proposed Method}

\definecolor{codeboxcolor}{RGB}{0, 73, 255}
\definecolor{codeboxcolorl}{RGB}{0, 73, 255}
\newtcolorbox{codebox}[1]{colback=codeboxcolorl!5!white,colframe=codeboxcolor,fonttitle=\bfseries,title=#1,left=-0.5mm,right=-0.1mm,top=0mm,bottom=0mm,toptitle=-0.2mm,bottomtitle=-0.8mm}

We introduce DISCODE, a novel test-time adaptive decoder that improves the robustness of LVLM-based image captioning evaluation.
By minimizing the ATT loss, which measures divergence between the token probability distribution and a Gaussian prior distribution, DISCODE adaptively generates robust scores at test time.
To the best of our knowledge, we are the first to propose a test-time adaptive approach for LVLM-based image captioning evaluation.

\subsection{Overview}

\paragraph{Problem Setting}
The goal of image captioning evaluation is to assign real-valued scores to image-caption pairs that better align with human judgments.
We assume the availability of a pre-trained LVLM that, given an appropriate prompt, produces raw evaluation scores $s_{\text{raw}}$ in $S = \{0, 1, \cdots, 9\}$.
We denote by $p_{\LVLM}$ the output token probability distribution over $S$, extracted at the token index $T$ corresponding to $s_{\text{raw}}$.
For example, if the LVLM output is ``Score: 7'', then $T$ indicates the index of the digit 7, and $p_{\LVLM}(7)$ represents the model's confidence in this score.
Thus, the raw score $s_{\text{raw}}$ is the digit $s \in S$ with the highest probability $p_{\LVLM}(s)$.

\paragraph{Scoring Pipeline}
Figure~\ref{fig:discode}{~(a)} shows the scoring pipeline of DISCODE, which consists of three steps.
First, given an input image and a candidate caption, the LVLM is prompted to generate a raw score $s_{\text{raw}} \in S$.
In this step, we extract the latent feature $\bm{h}_{T} \in \mathbb{R}^{d}$ from the last decoder layer of the LVLM along with the token probability distribution $p_{\LVLM}$.
Second, a probability mass distribution $p: S \to [0, 1]$ is estimated, which we refer to as the \textit{score distribution}.
This is a core step of the proposed pipeline, where DISCODE generates $p$ by minimizing the ATT loss.
Third, the final evaluation score $s$ is computed as the expected value, \textit{i.e.}, $s = \mathbb{E}_{x \sim p(x)}[x]$, thus yielding a real-valued score.

\paragraph{DISCODE}
DISCODE $\psi_{\theta}$ is a learnable decoder head to generate the score distribution $p$ from the latent decoder feature $\bm{h}_{T}$ as $p = \psi_{\hat{\theta}} (\bm{h}_{T})$.
The parameter $\theta$ is determined by minimizing the ATT loss:
\begin{align}
\label{eq:minimization_problem}
\hat{\theta} = \argmin_{\theta} \mathcal{L}_{\ATT} (\theta; \bm{h}_{T}).
\end{align}
This minimization problem is solved independently at test time for each image-caption pair, thereby enabling test-time adaptation across diverse visual domains.

\subsection{Adaptive Test-Time Loss}

The ATT loss is defined over two probability distributions: a prior distribution $q$ and the token probability distribution $p_{\LVLM}$.
As shown in Figure~\ref{fig:discode}(b), minimizing the ATT loss optimizes the balance between these two distributions, resulting in a robust estimation of the score distribution.

\paragraph{Loss Definition}
The ATT loss $\mathcal{L}_{\text{\scalebox{0.8}{ATT}}}$ consists of two terms, the cross-entropy term and the divergence term:
\begin{align}
\label{eq:att}
\mathcal{L}_{\text{\scalebox{0.8}{ATT}}}(\theta; \bm{h}_{T})
&= \underbrace{H(\psi_{\theta}(\bm{h}_{T}), p_{\LVLM})}_{\text{cross-entropy}} 
+ \underbrace{D_{\alpha} (\psi_{\theta}(\bm{h}_{T}) \| q)}_{\text{divergence}},
\end{align}
where $H$ is the cross entropy and $D_{\alpha}$ is the divergence measure.
As minimizing the cross-entropy term reduces the discrepancy between the estimated score distribution $\psi_{\theta} (\bm{h}_{T})$ and $p_{\LVLM}$, the divergence term can be understood as a regularization term to improve the robustness.

\paragraph{Prior\hspace{2pt}Distribution\hspace{2pt}$\bm{q}$}
Due to the central limit theorem, human evaluation scores naturally tend to follow a Gaussian distribution.
To reflect this, we use a Gaussian prior distribution $q(x) \propto \exp\left(-(x - s_{\text{raw}})^2/2\right)$, where $s_{\text{raw}} \in S$ is the raw score generated by the LVLM.

\paragraph{Divergence Term $\bm{D_{\alpha}}$}
In numerical evaluations, the highest and lowest scores are often easier to assess than intermediate ones. Consequently, when LVLMs predict scores near the minimum or maximum values, we can rely more on the raw scores and mitigate symbolic bias strongly by assigning greater weight to the unimodal prior.
To account for this,
we introduce the weighted Kullback-Leibler (KL) divergence for the divergence term, which adaptively determines the dependency on the prior with a parameter $\alpha \in [0, 1]$. Specifically, we define the divergence term by
\begin{align}
\label{eq:wkld}
D_{\alpha} (p \| q) = (1-\alpha) H(p, q)  - \alpha H(p, p),
\end{align}
where $H(p, q) = - \sum_{x \in S} p(x) \log q(x)$ is the cross-entropy.
The parameter $\alpha$ is adaptively determined based on the raw score using a Gaussian distribution as follows:
\begin{align}
\label{eq:alpha}
\alpha
=
\frac{1}{\sqrt{2\pi\sigma^2}} \exp\left(-\frac{(s_{\text{raw}} - \mu)^2}{2\sigma^2}\right)
\end{align}
where $\mu = |S|^{-1} \sum_{x \in S} x$ is the mean over the candidate digits and $\sigma^{2} = 0.1$ is a variance.
Note that $D_{\alpha}$ is equivalent to the vanilla Kullback-Leibler divergence when $\alpha = 0.5$.

\paragraph{Architecture for $\bm{\psi_{\theta}}$}
We employ a simple yet effective architecture for DISCODE $\psi_{\theta}$, which consists of a linear layer and a softmax function:
\begin{align}
\label{eq:arc}
\psi_{\theta}(\bm{h}) = \mathrm{softmax} (W^{\top} \bm{h} + \bm{b})
\end{align}
where $\theta = \{W, \bm{b}\}$ is a set of parameters, $W \in \mathbb{R}^{d \times |S|}$ is a weight matrix, $b \in \mathbb{R}^{|S|}$ is a bias vector.

\subsection{Analytical Solution}
Numerically solving the loss minimization problem in Eq.~(\ref{eq:minimization_problem}) is computationally expensive.
This limitation can be theoretically addressed by deriving an analytical solution.
Specifically, the analytical solution of the minimization problem exists for the loss defined in Eq.~(\ref{eq:att}) and the architecture defined in Eq.~(\ref{eq:arc}) under the assumption that the LVLM has a linear projection head that predicts token probabilities as $p_{\LVLM} = \mathrm{softmax} (V^{\top} \bm{h}_{T} + \bm{c})$. The solution is given by $\hat{\theta} = \{ \hat{W}, \hat{\bm{b}}\}$ with
\begin{align}
\label{tab:analytic}
\hat{W} = \frac{1}{\alpha} V,
\quad
\hat{\bm{b}} = \frac{1-\alpha}{\alpha} \log \bm{q} + \frac{1}{\alpha} \bm{c},
\end{align}
where $\bm{q} \in \mathbb{R}^{10}$ is the vector representation of the prior distribution $q$. The log function is applied element-wise.
The proof is provided in Appendix A.

\begin{figure}[t]
\centering
\includegraphics[width=\linewidth]{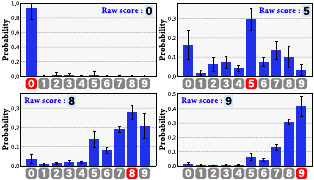}
\caption{Observed output token probability distributions for four target digits (0, 5, 8 and 9) with LLaVA-Next.}
\label{fig:observation}
\end{figure}

\subsection{Implementation Details}
\label{sec:implementation}
\begin{figure}
\centering
\includegraphics[width=\linewidth]{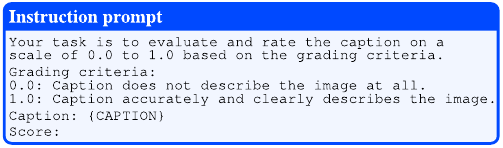}
\caption{Instruction prompt.}
\label{fig:prompt}
\end{figure}

Figure~\ref{fig:prompt} shows the default instruction prompt used in our experiments. This prompt is designed in the FLEUR framework~\cite{Lee2024FLEUR} to produce raw scores on a scale from 0.0 to 1.0.
For this prompt, the target is the first decimal place of the raw score. After computing the smoothed score $\hat{s}$, the value below the decimal point in the raw score is replaced with $0.1 \times \hat{s}$ as shown in Figure~\ref{fig:discode}{~(d)}.
We build DISCODE on top of ten open-source LVLMs:
LLaVA-Next-8B, -13B, -34B, -72B~\cite{li2024llavanext-strong},
InternVL-2.5-8B, -78B~\cite{Chen2024InternVL2_5},
Qwen2-VL-Instruct-7B, -72B~\cite{Wang2024Qwen2VL},
CogVLM2-Chat~\cite{Hong2024CogVLM2} and 
MiniCPM-V-2.6~\cite{Yao2024MiniCPMv}.
LLaVA-Next-72B is used as a default LVLM.
When using divergences other than KLD for an ablation study, we minimized the loss using the Adam optimizer for ten iterations, with an initial learning rate set to $10^{-3}$, because an analytical solution is not available.

\subsection{Discussion: Why is DISCODE effective?}
\label{sec:motivation}

Previous studies~\cite{Liu2023GEval, Lee2024FLEUR, Tong2025GVEval} assumed that the score distribution $p$ is identical to the output token probability distribution $p_{\LVLM}$ assigned by the LVLM, \textit{i.e.}, $p = p_{\LVLM}$.
However, $p_{\LVLM}$ is not necessarily aligned with the human judgment score distribution, particularly with respect to unimodality.
When a sufficiently large number of human evaluators provide scores, the resulting score distribution typically follows a unimodal distribution.\footnotemark
In contrast, the token probability distribution may be influenced by unintended bias such as symbolic bias.
In fact, the probability of the digit 0 is often overestimated as shown in Figure~\ref{fig:observation}, making the distribution non-unimodal.
DISCODE is effective because it addresses this issue by robustly estimating the score distribution through minimization of the ATT loss using a Gaussian prior distribution.
\footnotetext{More precisely, human evaluations may become polarized into two opposing extremes, resulting in a convex but not unimodal distribution. However, such cases rarely occur within the scope of quality evaluation.}

\begin{table}[t]
\centering
\small
\begin{tabular}{lccc}
\toprule
Dataset & Domain & Size & Avg length\\
\midrule
Flickr8k-Expert & Real & 5,664 & 11.91\\ 
Flickr8k-CF & Real & 47,830	& 11.35\\
Composite & Real & 13,146 & 11.58\\
Pascal-50S & Real & 8,000 & 8.75\\
MCEval & \textbf{6 domains} & 12,000 & 13.61\\
\bottomrule
\end{tabular}
\caption{Comparison with representative image captioning evaluation datasets. Size indicates the number of candidate captions to be evaluated.
}
\label{tab:dataset_statistics}
\end{table}
\section{MCEval Benchmark}

\begin{figure*}
\centering
\includegraphics[width=\linewidth]{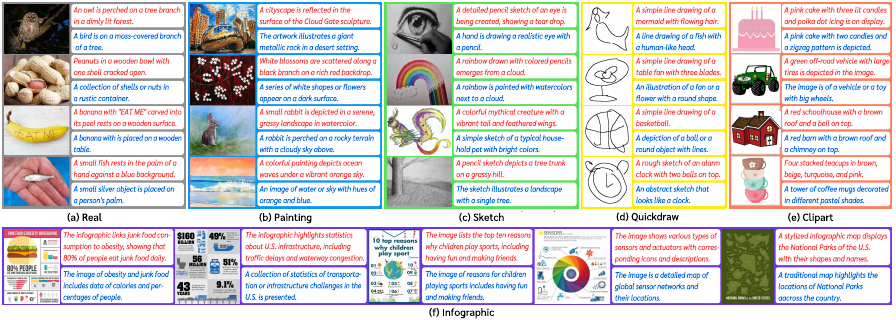}
\caption{MCEval dataset for benchmarking domain generalization in caption evaluation. Each image is paired with two candidate captions and a human preference label. Preferred captions are highlighted in red.}
\label{fig:rice}
\end{figure*}

\definecolor{lightcyan}{rgb}{0.88,0.95,1}
\begin{table*}[t]
\small
\centering
\setlength{\tabcolsep}{3.2pt}
\begin{tabular}{clcccccccccccccccc}
\toprule
& \textbf{\hspace{-12pt}Metric} & \textbf{LVLM} & \textbf{FF}
& & \textbf{Real}
& & \textbf{Painting}
& & \textbf{Sketch}
& & \textbf{Quickdraw}
& & \textbf{Clipart}
& & \textbf{Infograph}
& & \textbf{Mean}\\
\midrule
\multirow{5}{*}{\rotatebox{90}{\textbf{Reference-based}}}
& BLEU-4~\cite{Papineni2002BLEU} & & --  && 59.3 && 58.1 && 61.5 && 55.7 && 54.9 && 53.2 && 57.1 \\
& ROUGE~\cite{Lin2004ROUGE} & & -- && 57.9 && 56.9 && 59.5 && 54.5 && 54.4 && 50.6 && 55.6 \\
& METEOR~\cite{Banerjee2005METEOR} & & -- && 67.8 && 64.8 && 67.5 && 61.8 && 60.8 && 59.4 && 63.7 \\
& CIDEr~(Vedantam et al. 2015) & & --  && 66.7 && 64.5 && 68.7 && 62.8 && 64.5 && 60.2 && 64.6 \\
\cmidrule{2-18}
& BERT-S~\cite{Zhan2020BERTScore} & &$\checkmark$&& 68.5 && 73.8 && 74.3 && 72.6 && 67.0 && 58.6 && 69.1 \\
& Polos~\cite{Wada2024PolosPolaris} &&&& 81.3 && 75.0 && 77.6 && 76.8 && 74.5 && 69.0 && 75.7 \\
\midrule
\multirow{6}{*}{\rotatebox{90}{\textbf{Reference-free~~}}}
& CLIP-S~\cite{Hessel2021CLIPS} & & $\checkmark$ && 79.2 && 78.0 && 78.3 && 75.4 && 73.9 && 66.7 && 75.3 \\
& PAC-S~\cite{Sarto2023PACS} &&&& 80.7 && 71.1 && 69.7 && 67.5 && 66.8 && 58.7 && 69.1 \\
\cmidrule{2-18}
& FLEUR  (Lee et al. 2024) & $\checkmark$ & $\checkmark$ && 84.7 && 83.6 && 80.4 && 45.6 && 79.9 && 86.0 && 76.7\\
& G-VEval (GPT-4o)~\cite{Tong2025GVEval} & $\checkmark$ & $\checkmark$ && 
86.0 && 80.2 && 81.2 && 76.9 && 80.6 &&  81.0 && 81.0\\
& FLEUR$^{\dagger}$ & $\checkmark$ & $\checkmark$ && 86.9 && 84.3 && 83.1 && 76.3 && 82.0 && 82.3 && 82.5\\
& \cellcolor{lightcyan}\textbf{DISCODE (Ours)} & \cellcolor{lightcyan}$\checkmark$ & \cellcolor{lightcyan}$\checkmark$ &\cellcolor{lightcyan}& \cellcolor{lightcyan}\textbf{87.8} &\cellcolor{lightcyan}& \cellcolor{lightcyan}\textbf{85.2} &\cellcolor{lightcyan}& \cellcolor{lightcyan}\textbf{83.9} &\cellcolor{lightcyan}& \cellcolor{lightcyan}\textbf{78.5} &\cellcolor{lightcyan}& \cellcolor{lightcyan}\textbf{83.5} &\cellcolor{lightcyan}& \cellcolor{lightcyan}\textbf{82.8} &\cellcolor{lightcyan}& \cellcolor{lightcyan}\textbf{83.6}\\[-0.2em]
\bottomrule
\end{tabular}
\caption{
Performance comparison on the MCEval benchmark.
Marks for FF indicate finetuning-free metrics.
$^{\dagger}$ indicates our implementation of FLEUR using the same LVLM (LLaVA-Next-72B) as DISCODE for a fair comparison.
}
\label{tab:rice}
\end{table*}

This section presents MCEval, a novel human evaluation dataset for benchmarking the robustness and generalizability of evaluation metrics.
MCEval consists of 6,000 images, each with two candidate captions and one reference caption, totaling 18,000 image-caption pairs across six domains.

\paragraph{Dataset Construction}
We constructed MCEval using images from DomainNet~\cite{peng2019domainnet} and InfographicVQA~\cite{Mathew2022InfographicVQA}, covering six visual domains: \textit{real}, \textit{painting}, \textit{sketch}, \textit{quickdraw}, \textit{clipart}, and \textit{infograph}.
Our  annotation process involved three steps.
First, we randomly sampled 1,000 images for each domain and generated their candidate captions using four proprietary LVLMs: GPT-4o-mini, GPT-4o, Gemini 2.0 Flash, and Claude 3.5 Sonnet.
Open-source LVLMs were excluded because DISCODE depends on them.
Second, we randomly selected one candidate caption per image, and human annotators revised it to form a reference caption.
Third, for each image, three annotators compared two other candidate captions, selecting the better caption based on relevance, descriptiveness, correctness, and fluency. If consensus among these annotators was not reached, the image was discarded and replaced with another randomly sampled image from the same domain. Annotation tasks were completed by 81 crowdworkers via MTurk and Upwork platforms.

\paragraph{Dataset Statistics}
Table~\ref{tab:dataset_statistics} compares MCEval with other widely-used image captioning evaluation datasets in terms of domain coverage, number of candidate captions, and average caption length (words per caption).
As shown, MCEval contains a moderate number of samples for benchmarking performance and covers a broader range of domains compared to existing datasets.

\paragraph{Examples} Figure~\ref{fig:rice} shows example images, each paired with two candidate captions. As shown, MCEval presents challenging scenarios because it involves images that are more abstract than realistic, accompanied by captions focusing on shapes and visual patterns.

\def\rawscorellavanextseventytwo{%
\flkcom{31.9}{32.2}{35.5}{18.4}{58.3}{63.1}
\pascalraw{31.4}{99.6}{96.6}{41.7}{67.3}
\mdce{61.6}
}%
\def\smoothingllavanextseventytwo{%
\flkcom{55.3}{55.7}{40.1}{20.7}{60.7}{65.7}
\pascal{61.2}{99.3}{96.5}{\textbf{77.7}}{83.8}
\mdce{82.5}
}%
\def\discodellavanextseventytwo{%
\flkcomb{55.7}{56.1}{40.2}{20.8}{61.1}{66.0}
\pascal{\textbf{62.9}}{\textbf{99.8}}{\textbf{97.9}}{77.4}{\textbf{84.5}}
\mdce{\textbf{83.6}}
}%
\def\discodellavanextseventytwox{%
\flkcomb{55.7}{56.1}{40.2}{20.8}{61.1}{66.0}
\pascal{62.9}{\textbf{99.8}}{97.9}{\textbf{77.4}}{\textbf{84.5}}
\mdce{\textbf{83.6}}
}%
\def\discodellavanextseventytwoy{%
\flkcom{55.7}{\textbf{56.1}}{\textbf{40.2}}{20.8}{61.1}{\textbf{66.0}}
\pascal{62.9}{99.8}{97.9}{77.4}{\textbf{84.5}}
\mdce{\textbf{83.6}}
}
\newif\ifhalfsize

\halfsizefalse
\def\multirowforfull#1{\multirow{2}{*}{#1}}
\def\disableforhalf#1{#1}
\def\tablebegin{\begin{table*}[t]}
\def\tableend{\end{table*}}
\def\tabularbegin{\begin{tabular}{l c cc c cc c cc c cc c ccccc c cc}}
\def\tabularend{\end{tabular}}
\def\datasetlist{ \multicolumn{2}{c}{\textbf{Flickr8k-Expert}} & & \multicolumn{2}{c}{\textbf{Flickr8k-CF}} & & \multicolumn{2}{c}{\textbf{Composite}} & & \multicolumn{5}{c}{\textbf{Pascal-50S}}\\
\cmidrule{3-4} \cmidrule{6-7} \cmidrule{9-10} \cmidrule{12-16}
& & \scalebox{0.9}{Kendall\hspace{1pt}$\tau_b$} & \scalebox{0.9}{Kendall\hspace{1pt}$\tau_c$} & & \scalebox{0.9}{Kendall\hspace{1pt}$\tau_b$} & \scalebox{0.9}{Kendall\hspace{1pt}$\tau_c$} & & \scalebox{0.9}{Kendall\hspace{1pt}$\tau_b$} & \scalebox{0.9}{Kendall\hspace{1pt}$\tau_c$} & & \scalebox{0.9}{HC} & \scalebox{0.9}{HI} & \scalebox{0.9}{HM} & \scalebox{0.9}{MM} & \scalebox{0.9}{Mean}
}

\def\flkcom#1#2#3#4#5#6{#1 & #2 & & #3 & #4 & & #5 & #6 & &}
\def\pascal#1#2#3#4#5{#1 & #2 & #3 & #4 & #5}
\def\pascalraw#1#2#3#4#5{#1 & #2 & #3 & #4 & #5}
\def\rice#1#2{#1 & #2 & &}
\def\mdce#1{}

\def\flkcomb#1#2#3#4#5#6{\flkcom{\textbf{#1}}{\textbf{#2}}{\textbf{#3}}{\textbf{#4}}{\textbf{#5}}{\textbf{#6}}}
\def\pascalb#1#2#3#4#5{\pascal{\textbf{#1}}{\textbf{#2}}{\textbf{#3}}{\textbf{#4}}{\textbf{#5}}}

\def\Smoothing{FLEUR}
\def\learningbased{}

\def\result#1#2#3#4{#1 & & #2 & & #3 & & #4}

\begin{table*}[t]
\small
\centering
\setlength{\tabcolsep}{1pt}
\begin{tabular}{c l c cc c cc c cc c c}
\toprule
& \hspace{-10pt}\multirow{2}{*}{\vspace{-4pt}\textbf{Metric}}
& \multirow{2}{*}{\hspace{-5pt}\vspace{-4pt}\textbf{LVLM}\hspace{3pt}}
& \multicolumn{2}{c}{\textbf{Flickr8k-Expert}} & 
& \multicolumn{2}{c}{\textbf{Flickr8k-CF}} & 
& \multicolumn{2}{c}{\textbf{Composite}} & 
& \textbf{Pascal-50S} \\
\cmidrule{4-5}\cmidrule{7-8}\cmidrule{10-11}\cmidrule{13-13}
&  &  & Kendall $\tau_b$ & Kendall $\tau_c$
&  & Kendall $\tau_b$ & Kendall $\tau_c$
&  & Kendall $\tau_b$ & Kendall $\tau_c$
&  & Accuracy \\
\midrule
\multirow{9}{*}{\rotatebox{90}{\textbf{Reference-based}}}
& BLEU-4~\cite{Papineni2002BLEU} &                     & 30.6 & 30.8 & & 16.9 &  8.7 & & 28.3 & 30.6 & & 74.0 \\
& ROUGE~\cite{Lin2004ROUGE}     &                     & 32.1 & 32.3 & & 19.9 & 10.3 & & 30.0 & 32.4 & & 78.0 \\
& METEOR~\cite{Banerjee2005METEOR} &                  & 41.5 & 41.8 & & 22.2 & 11.5 & & 36.0 & 38.9 & & 81.1 \\
& CIDEr~\cite{Vedantam2015CIDEr}&                     & 43.6 & 43.9 & & 24.6 & 12.7 & & 34.9 & 37.7 & & 80.1 \\
& SPICE~\cite{Anderson2016SPICE}&                     & 51.7 & 44.9 & & 24.4 & 12.0 & & 38.8 & 40.3 & & 78.7 \\
\cmidrule{2-13}
& RefCLIP-S~\cite{Hessel2021CLIPS}              &                     & 52.6 & 53.0 & & 36.4 & 18.8 & & 51.2 & 55.4 & & 83.3 \\
& RefPAC-S++ (ViT-L)~\cite{Sarto2024PACScorepp} &                     & --   & 57.9 & & 38.8 &   -- & &   -- & 61.6 & & 84.7 \\
& Polos~\cite{Wada2024PolosPolaris}             &                     & --   & 56.4 & & 37.8 &   -- & &   -- & 57.6 & & 86.5 \\
& DENEB (ViT-L)~(Matsuda et al. 2024) &                     & --   & 56.8 & & 38.3 &   -- & &   -- & 58.2 & & \textbf{87.8} \\
\midrule
\multirow{14}{*}{\rotatebox{90}{\textbf{Reference‑free}}}
& UMIC~\cite{Lee2021UMIC}                 & \learningbased      &  --  & 46.8 & &  --  &  --  & &  --  & 56.1 & & 85.1 \\
& CLIP-S~\cite{Hessel2021CLIPS}           &                     & 51.1 & 51.2 & & 34.4 & 17.7 & & 49.8 & 53.8 & & 80.9 \\
& InfoMetIC+~\cite{Hu2023InfoMetIC}       &                     &  --  & 55.5 & & 36.6 &  --  & &  --  & 59.3 & & 86.5 \\
& HiFi‑Score~\cite{yao2024hifiscore}      &                     &  --  & 58.4 & &  --  &  --  & &  --  & 65.8 & & 83.0 \\
& BRIDGE (ViT‑L)~\cite{Sarto2024BRIDGE}   &                     & 55.4 & 55.8 & & 36.3 & 19.0 & & 52.9 & 57.2 & & 82.9 \\
& HICE‑S~\cite{zeng2024hicescore}         &                     & 55.9 & 56.4 & & 37.2 & 19.2 & & 53.1 & 57.9 & & 86.1 \\
& PAC‑S++ (ViT‑L)~\cite{Sarto2024PACScorepp}&                   &  --  & 57.4 & & 38.5 &  --  & &  --  & 62.0 & & 82.4 \\
& CAMScore~\cite{cui2025evaluating}       &                     & 54.8 & 55.6 & & 37.5 & 19.3 & & 53.4 & 57.5 & & 85.8 \\
\cmidrule{2-13}
& FLEUR~(Lee et al. 2024)   & $\checkmark$        &  --  & 53.0 & & 38.6 &  --  & &  --  & 63.5 & & 83.2 \\
& G‑VEval (GPT‑4o)~\cite{Tong2025GVEval}  & $\checkmark$        & \underline{\textbf{61.5}} & \underline{\textbf{59.7}}
 & & 38.7 & 20.2 & & 58.3 & 63.0 & & 82.3 \\
& FLEUR‑LV$^{\dagger}$                    & $\checkmark$        & 55.3 & 55.7 & & 40.1 & 20.7 & & 60.7 & 65.7 & & 83.8 \\
& \cellcolor{lightcyan}\textbf{DISCODE‑LV (Ours)}
 & \cellcolor{lightcyan}$\checkmark$
 & \cellcolor{lightcyan}55.7 & \cellcolor{lightcyan}56.1
 & \cellcolor{lightcyan}& \cellcolor{lightcyan}\underline{\textbf{40.2}}
 & \cellcolor{lightcyan}\underline{\textbf{20.8}}
 & \cellcolor{lightcyan}& \cellcolor{lightcyan}\underline{\textbf{61.1}}
 & \cellcolor{lightcyan}\underline{\textbf{66.0}}
 & \cellcolor{lightcyan}& \cellcolor{lightcyan}\underline{84.5} \\
& FLEUR‑IN$^{\dagger}$                    & $\checkmark$        & 56.5 & 56.9 & & 36.4 & 18.8 & & 59.8 & 64.2 & & 80.8 \\
& \cellcolor{lightcyan}\textbf{DISCODE‑IN (Ours)}
 & \cellcolor{lightcyan}$\checkmark$
 & \cellcolor{lightcyan}57.7 & \cellcolor{lightcyan}58.1
 & \cellcolor{lightcyan}& \cellcolor{lightcyan}40.1
 & \cellcolor{lightcyan}\underline{\textbf{20.8}}
 & \cellcolor{lightcyan}& \cellcolor{lightcyan}60.5
 & \cellcolor{lightcyan}64.9
 & \cellcolor{lightcyan}& \cellcolor{lightcyan}83.5 \\[-0.25em]
\bottomrule
\end{tabular}
\caption{Comparison with state-of-the-art metrics on Flickr8k-Expert, Flickr8k-CF, Composite and Pascal-50S.
FLEUR$^{\dagger}$ and DISCODE utilize LLaVA-Next-72B (LV) and InternVL-2.5-78B (IN).
Best results are marked in bold, and best results among LVLM-based approaches are underlined.
}
\label{tab:sota_main}
\end{table*}

\section{Experiments}

\subsection{Experimental setting}

\paragraph{Datasets and performance measures}
We conducted extensive experiments to demonstrate the effectiveness of DISCODE on MCEval and four commonly used benchmarks: Flickr8k-Expert~\cite{Hodosh2013Flickr8kEXCF}, Flickr8k-CF~\cite{Hodosh2013Flickr8kEXCF}, Composite~\cite{Aditya2015Composite} and Pascal-50S~\cite{Vedantam2015CIDEr}.
For MCEval, we report accuracy for each visual domain as well as the mean accuracy.
For Flickr8k-Expert, Flickr8k-CF, and Composite, we utilize Kendall’s tau-b ($\tau_{b}$) and tau-c ($\tau_{c}$) as performance measures.
For Pascal-50S, we report mean accuracy over the HC, HI, HM, and MM annotation types.

\paragraph{Baselines} We compare DISCODE with eight competitive baselines on MCEval, including three reference-free metrics: CLIP-S~\cite{Hessel2021CLIPS}, PAC-S~\cite{Sarto2023PACS} and FLEUR \cite{Lee2024FLEUR}, and five reference-based
metrics: BLEU~\cite{Papineni2002BLEU}, ROUGE~\cite{Lin2004ROUGE}, METEOR~\cite{Banerjee2005METEOR}, CIDEr~\cite{Vedantam2015CIDEr}, 
and Polos~\cite{Wada2024PolosPolaris}.
FLEUR is the most relevant to our work, as it uses score smoothing.

\paragraph{Implementation details}
DISCODE is implemented as described in Section~\ref{sec:implementation}.
For learning-based metrics, we utilize their officially released pre-trained checkpoints.

\definecolor{LightCyan}{rgb}{0.88,0.95,1}

\subsection{Experimental results}

\paragraph{Performance\hspace{3pt}on\hspace{3pt}Diverse\hspace{3pt}Domains}
Table~\ref{tab:rice} summarizes the performance comparison on the MCEval benchmark.
Overall, DISCODE consistently achieves superior performance across all visual domains, highlighting its robustness to diverse image styles such as paintings and abstract drawings.
We observe performance drops in learning-based methods like PAC-S and Polos on non-real domains, likely due to their finetuning targeting real images.
By contrast, DISCODE exhibits greater stability across domains, outperforming FLEUR due to its more robust score estimation strategy via the ATT loss minimization. 

\paragraph{Performance\hspace{3pt}on\hspace{3pt}Real\hspace{3pt}Image\hspace{3pt}Domain}
Table~\ref{tab:sota_main}
compares DISCODE with state-of-the-art methods on the four real-image benchmark datasets.
DISCODE achieves comparable or even better performance than the other methods.
On Flickr8k-CF, it outperforms G-VEval, which utilizes a proprietary LVLM (GPT-4o), demonstrating the strong adaptability of DISCODE in real-image evaluation scenarios.

\subsection{Analysis}

\paragraph{Ablation Study}
Table~\ref{tab:ablation} presents the ablation study results examining three key DISCODE components:
the cross-entropy term $H$ in Eq.~\eqref{eq:att},
the divergence term $D_{\alpha}$ in Eq.~\eqref{eq:att},
and the weighting parameter $\alpha$ in Eq.~\eqref{eq:alpha}.
The performance drop resulting from the removal of each component confirms their collective contribution to the model.

\def\disableforhalf#1{}
\def\disableforhalfx#1{#1}
\def\multirowforfull#1{#1}

\halfsizetrue

\def\tablebegin{\begin{table}[t]}
\def\tableend{\end{table}}
\def\tabularbegin{\begin{tabular}{l cc cc c cc c cc c ccccc c cc}}
\def\tabularend{\end{tabular}}
\def\datasetlist{FEX\hspace{1pt}$\tau_{c}$ & FCF\hspace{1pt}$\tau_{b}$ & Com\hspace{1pt}$\tau_{c}$ & Pascal & MCEval}
\def\mdce#1{& #1}
\def\rice#1#2{#2 &}
\def\flkcom#1#2#3#4#5#6{#2 & #3 & #6 & }
\def\pascal#1#2#3#4#5{#5}
\def\pascalraw#1#2#3#4#5{#5}
\def\rice#1#2{#2 &}
\def\flkcomb#1#2#3#4#5#6{\flkcom{\textbf{#1}}{\textbf{#2}}{\textbf{#3}}{\textbf{#4}}{\textbf{#5}}{\textbf{#6}}}
\def\pascalb#1#2#3#4#5{\pascal{\textbf{#1}}{\textbf{#2}}{\textbf{#3}}{\textbf{#4}}{\textbf{#5}}}

{{\setlength{\tabcolsep}{.7pt}\def\textbfforhalf#1{
\ifhalfsize 
\textbf{#1}
\else
#1
\fi
}
\def\textbfforfull#1{\ifhalfsize#1\else\textbf{#1}\fi}
\tablebegin
\small
\centering
\resizebox{\linewidth}{!}{
\tabularbegin
\toprule
\disableforhalf{}
\multirowforfull{Method}
\disableforhalf{&}
&
\datasetlist\\
\midrule
\rowcolor{lightcyan}
DISCODE &
\disableforhalf{&}
\discodellavanextseventytwox
\\
w/o
cross-entropy term $H$ &
\disableforhalf{&}
\flkcom{54.3}{54.6}{39.5}{20.4}{58.3}{63.1}
\pascal{62.5}{\textbf{99.8}}{97.2}{75.7}{83.8}
\mdce{81.8}
\\
w/o 
divergence term $D_{\alpha}$
\disableforhalf{&}
&
\flkcom{49.6}{49.9}{39.9}{20.6}{59.6}{64.4}
\pascal{61.7}{99.7}{97.7}{72.7}{83.0}
\mdce{80.9}
\\
w/o 
adaptive definition for $\alpha$ &
\disableforhalf{&}
\flkcom{55.2}{55.6}{\textbf{40.2}}{\textbf{20.8}}{60.5}{65.4}
\pascal{\textbf{63.9}}{\textbf{99.8}}{\textbf{98.1}}{75.5}{84.3}
\mdce{83.0}
\\
\bottomrule
\tabularend
}
\ifhalfsize
\caption{
Ablation study results.
}
\label{tab:ablation}
\else
\caption{Full results for ablation study.
First two experiments omit the cross-entropy and divergence terms in Eq.~\eqref{eq:att}, respectively. Third experiment utilizes $\alpha=0.5$ to omit the adaptive definition for $\alpha$.\vspace{1em}}
\label{tab:ablation_appendix}
\small
\centering
\tabularbegin
\toprule
\multirowforfull{\textbf{Variance} $\sigma^{2}$} & & \datasetlist \\
\midrule
1.0 & &
\flkcom{55.6}{55.9}{40.2}{20.8}{60.9}{65.9}
\pascal{61.8}{99.8}{97.9}{77.2}{84.2}\\
0.5 & &
\flkcom{55.6}{55.9}{40.2}{20.8}{60.9}{65.9}
\pascal{61.8}{99.8}{97.9}{77.2}{84.2}\\
\rowcolor{lightcyan} 0.1 & &
\flkcom{55.7}{56.1}{40.2}{20.8}{61.1}{66.0}
\pascal{62.9}{99.8}{97.9}{77.4}{84.5}\\
0.05 & &
\flkcom{55.7}{56.1}{40.2}{20.8}{61.1}{66.0}
\pascal{62.9}{99.7}{97.9}{77.4}{84.5}\\
0.01 & &
\flkcom{55.7}{56.1}{40.2}{20.8}{61.1}{66.0}
\pascal{63.0}{99.7}{97.9}{76.5}{84.3}\\
\bottomrule
\tabularend
\caption{Hyperparameter study for $\sigma$ to adaptively define $\alpha$.}
\label{tab:sigma}
\fi
\tableend
}
{\setlength{\tabcolsep}{5.2pt}\tablebegin
\centering
\small
\setlength{\tabcolsep}{6.2pt} 
\begin{tabular}{ll c c c c c c c c c c}
\toprule
\disableforhalf{\multirowforfull{Method}}
\ifhalfsize \else & \fi
\multirowforfull{Scale}
&
\datasetlist \\
\midrule
\disableforhalf{
Raw score & 1 to 5 & 
\flkcom{22.3}{22.5}{30.0}{15.5}{51.2}{55.4}
\pascal{25.5}{99.1}{95.1}{23.4}{60.7}
\mdce{\red{WIP}}
\\
Raw score & 0 to 9 & 
\flkcom{21.3}{21.4}{29.1}{15.0}{55.6}{60.1}
\pascal{36.3}{98.9}{97.0}{32.1}{66.1}
\mdce{\red{WIP}}
\\
Raw score & A to E & 
\flkcom{29.0}{29.2}{33.2}{17.2}{53.3}{57.6}
\pascal{8.9}{99.4}{91.8}{31.3}{57.8}
\mdce{\red{WIP}}
\\
Raw score & 0.0 to 1.0 & 
\rawscorellavanextseventytwo
\\
\midrule
Smoothing & 1 to 5 & 
\flkcom{52.1}{52.5}{39.8}{20.5}{59.9}{64.7}
\pascal{62.0}{99.7}{98.0}{74.4}{83.5}
\mdce{\red{WIP}}
\\
Smoothing & 0 to 9 & 
\flkcom{54.0}{54.4}{40.1}{20.7}{60.7}{65.6}
\pascal{62.4}{\textbf{99.8}}{97.9}{76.6}{84.1}
\mdce{\red{WIP}}
\\
Smoothing & A to E & 
\flkcom{53.5}{53.8}{39.8}{20.6}{60.6}{65.5}
\pascal{61.4}{99.7}{97.5}{75.5}{83.5}
\mdce{\red{WIP}}
\\
\Smoothing & 0.0 to 1.0 & 
\smoothingllavanextseventytwo
\\
\midrule
}
\disableforhalf{DISCODE &}
1 to 5 &
\flkcom{54.0}{54.4}{40.1}{20.7}{60.2}{65.1}
\pascal{62.8}{99.7}{\textbf{98.1}}{77.2}{\textbf{84.5}}
\mdce{\textbf{83.6}}
\\
\disableforhalf{DISCODE &}
0 to 9 &
\flkcom{55.1}{55.4}{40.1}{20.7}{60.9}{65.8}
\pascal{\textbf{64.0}}{\textbf{99.8}}{97.7}{75.1}{84.2}
\mdce{83.5}
\\
\disableforhalf{DISCODE &}
A to E &
\flkcom{54.1}{54.4}{39.8}{20.6}{60.8}{65.7}
\pascal{61.8}{99.7}{97.4}{75.7}{83.7}
\mdce{83.5}
\\
\rowcolor{lightcyan}
\disableforhalf{DISCODE &} 0.0 to 1.0 &
\flkcomb{55.7}{56.1}{40.2}{20.8}{61.1}{66.0}
\pascal{62.9}{\textbf{99.8}}{97.9}{77.4}{\textbf{84.5}}
\mdce{\textbf{83.6}}
\\[-0.2em]
\bottomrule
\end{tabular}
\ifhalfsize
\caption{
Comparison of rating scales.
}
\label{tab:rating}
\else
\caption{Comparison of rating scales.}
\label{tab:rating_appendix}
\fi
\tableend

}}

\paragraph{Which rating scale is best?}
Table~\ref{tab:rating} compares four different rating scales:
(1) a continuous scale from 0.0 to 1.0,
(2) a five-point discrete scale from 1 to 5,
(3) a ten-point discrete scale from 0 to 9, and
(4) a letter-based scale from A to E.
We observe that numerical scales perform better, with the continuous scale slightly outperforming discrete scales. 
This confirms that instructing LVLMs to assign more granular scores is effective in improving correlation with human evaluations.
With the continuous scale, the decimal place token always appears immediately before the target digit. This likely helped stabilize the output token probability distribution during autoregressive decoding, resulting in improved performance compared to the ten-point discrete scale.

\paragraph{Generalization Across LLMs}
To investigate the compatibility of DISCODE with different LLMs, we applied it to four different LLMs within the framework of LLaVA-Next: Llama-3-8B~\cite{Dubey2024Llama3}, 
Vicuna-13B~\cite{vicuna2023}, Nous-Hermes-2-Yi-34B~\cite{NousResearch2024HermesYi} and Qwen-1.5-72B~\cite{QwenTeam2024qwen1.5}.
Results summarized in Table~\ref{tab:llms} indicate consistent improvements over raw scores and FLEUR in most cases, verifying the broad applicability of DISCODE.
We also see that larger models exhibit higher performance, highlighting the scalability advantages.

\paragraph{Generalization Across LVLMs}
To further explore the applicability of DISCODE,
we applied it to six leading LVLMs:
InternVL-2.5-8B/78B~\cite{Chen2024InternVL, Chen2024InternVL2_5},
Qwen2-VL-7B/72B-Instruct~\cite{Bai2023QwenVL, Wang2024Qwen2VL},
CogVLM2-Chat-19B~\cite{Wang2023CogVLM,Hong2024CogVLM2},
and Mini-CPM-V-2.6~\cite{Hu2024MiniCPM, Yao2024MiniCPMv}.
As shown in Table~\ref{tab:lmms}, 
DISCODE consistently improves performance across these models. 
Among all results in  Tables~\ref{tab:llms} and \ref{tab:lmms}, the best performance is achieved
by InternVL-2.5-78B on Flickr8k-Expert (58.1 $\tau_{c}$),
by Qwen-VL-72B-Instruct on Composite (66.7 $\tau_{c}$) and MCEval (83.8\%), 
and by LLaVA-Next-Qwen-1.5-72B on Flickr8k-CF (40.2 $\tau_{b}$) and Pascal-50S (84.5\%).
These results underscore that our approach is effective regardless of the pre-training methods or architectures of the LVLMs.
{\setlength{\tabcolsep}{3.7pt}\begin{table}[t]
\small
\centering
\begin{tabular}{ll *{12}{c}}
\toprule
LLM & \multirowforfull{Method}  & \datasetlist\\
\midrule
\multirow{3}{*}{\rotatebox[origin=c]{90}{\begin{minipage}{1cm}\centering LLama\\ 3-8B\end{minipage}}}
& Raw score & 21.3 & 23.6 & 47.5 & 57.2 & 52.1 \\
& \Smoothing & 49.6 & 33.5 & 50.4 & 79.2 & 76.8 \\
& \cellcolor{lightcyan}DISCODE &
\cellcolor{lightcyan} \textbf{51.1} &\cellcolor{lightcyan}  \textbf{36.2} &\cellcolor{lightcyan}  \textbf{61.2} &\cellcolor{lightcyan}  \textbf{81.9} &\cellcolor{lightcyan}  \textbf{77.4} \\[-0.2em]
\midrule
\multirow{3}{*}{\rotatebox[origin=c]{90}{\begin{minipage}{1cm}\centering Vicuna \\ 13B\end{minipage}}}
& Raw score & 24.5 & 29.8 & 57.0 & 64.4 & 55.0 \\
& \Smoothing & 52.1 & 38.0 & 61.9 & 83.5 & 80.6 \\
& \cellcolor{lightcyan}DISCODE & \textbf{\cellcolor{lightcyan}52.6} & \textbf{\cellcolor{lightcyan}38.4} & \textbf{\cellcolor{lightcyan}62.9} & \textbf{\cellcolor{lightcyan}83.9} & \textbf{\cellcolor{lightcyan}81.4} \\[-0.2em]
\midrule
\multirow{3}{*}{\rotatebox[origin=c]{90}{\begin{minipage}{1cm}\centering Hermes\\ Yi-34B\end{minipage}}}
& Raw score & 16.3 & 22.9 & 54.6 & 60.5 & 61.0 \\
& \Smoothing & 54.4 & \textbf{39.5} & \textbf{66.1} & 83.4 & 79.7 \\
& \cellcolor{lightcyan}DISCODE & \textbf{\cellcolor{lightcyan}55.0} & \textbf{\cellcolor{lightcyan}39.5} & \textbf{\cellcolor{lightcyan}66.1} & \textbf{\cellcolor{lightcyan}84.1} & \textbf{\cellcolor{lightcyan}82.5} \\[-0.2em]
\midrule
\multirow{3}{*}{\rotatebox[origin=c]{90}{\begin{minipage}{1.1cm}\centering Qwen\\ 1.5-72B\end{minipage}}} 
& Raw score & 32.2 & 35.5 & 63.1 & 67.3 & 61.6 \\
& \Smoothing & 55.7 & 40.1 & 65.7 & 83.8 & 82.5 \\
& \cellcolor{lightcyan}DISCODE & \textbf{\cellcolor{lightcyan}56.1} & \textbf{\cellcolor{lightcyan}40.2} & \textbf{\cellcolor{lightcyan}66.0} & \textbf{\cellcolor{lightcyan}84.5} & \textbf{\cellcolor{lightcyan}83.6} \\[-0.2em]
\bottomrule
\end{tabular}
\caption{Performance comparison across four LLMs within the LLaVA-NeXT framework.
}
\label{tab:llms}
\end{table}}
{\setlength{\tabcolsep}{.7pt}\begin{table}[t]
\small
\centering
\setlength{\tabcolsep}{3.1pt}
\begin{tabular}{cl *{12}{c}}
\toprule
LVLM & Metric & \datasetlist \\
\midrule
\multirow{3}{*}{\rotatebox[origin=c]{90}{\begin{minipage}{1cm}\centering IntVL \\ 8B\end{minipage}}} %
& Raw score & 26.9 & 30.4 & 55.0 & 66.0 & 57.9 \\
& \Smoothing & 56.3 & 36.7 & 58.4 & 76.0 & 64.5 \\
& \cellcolor{lightcyan}DISCODE & \textbf{\cellcolor{lightcyan}57.4} & \textbf{\cellcolor{lightcyan}39.2} & \textbf{\cellcolor{lightcyan}59.6} & \textbf{\cellcolor{lightcyan}80.2} & \textbf{\cellcolor{lightcyan}66.5} \\[-0.2em]
\midrule
\multirow{3}{*}{\rotatebox[origin=c]{90}{\begin{minipage}{1cm}\centering IntVL\\78B\end{minipage}}} 
& Raw score & 42.0 & 35.7 & 59.6 & 72.8 & 66.3 \\
& \Smoothing & 56.9 & 36.4 & 64.2 & 80.8 & 74.0 \\
& \cellcolor{lightcyan}DISCODE & \textbf{\cellcolor{lightcyan}58.1} & \textbf{\cellcolor{lightcyan}40.1} & \textbf{\cellcolor{lightcyan}64.9} & \textbf{\cellcolor{lightcyan}83.5} & \textbf{\cellcolor{lightcyan}78.1} \\[-0.2em]
\midrule
\multirow{3}{*}{\rotatebox[origin=c]{90}{\begin{minipage}{1cm}\centering Qwen2\\7B-I\end{minipage}}}
& Raw score & 14.5 & 39.1 & 51.7 & 60.1 & 53.3 \\[-0.2em]
 & \Smoothing & 52.6 & \textbf{39.5} & 52.7 & 81.5 & \textbf{75.2} \\
 & \cellcolor{lightcyan}DISCODE & \textbf{\cellcolor{lightcyan}52.9} & \textbf{\cellcolor{lightcyan}39.6} & \textbf{\cellcolor{lightcyan}66.2} & \textbf{\cellcolor{lightcyan}83.3} & \textbf{\cellcolor{lightcyan}75.2} \\[-0.2em]
\midrule
\multirow{3}{*}{\rotatebox[origin=c]{90}{\begin{minipage}{1cm}\centering Qwen2\\72B-I\end{minipage}}}
 & Raw score & 12.8 & 32.2 & 63.0 & 66.9 & 65.8 \\
 & \Smoothing & 54.1 & \textbf{40.0} & 66.3 & 83.7 & 82.4 \\
 & \cellcolor{lightcyan}DISCODE & \textbf{\cellcolor{lightcyan}54.4} & \textbf{\cellcolor{lightcyan}40.0} & \textbf{\cellcolor{lightcyan}66.7} & \textbf{\cellcolor{lightcyan}84.1} & \textbf{\cellcolor{lightcyan}83.8} \\[-0.2em]
\midrule
\multirow{3}{*}{\rotatebox[origin=c]{90}{\begin{minipage}{1cm}\centering CogVL\\19B\end{minipage}}}
& Raw score & 13.7 & 14.3 & 30.6 & 67.3 & 54.5 \\
 & \Smoothing & 39.1 & 29.8 & 44.2 & 79.6 & 66.1 \\
 & \cellcolor{lightcyan}DISCODE & \textbf{\cellcolor{lightcyan}40.3} & \textbf{\cellcolor{lightcyan}31.9} & \textbf{\cellcolor{lightcyan}53.0} & \textbf{\cellcolor{lightcyan}80.2} & \textbf{\cellcolor{lightcyan}68.6} \\[-0.2em]
\midrule
\multirow{3}{*}{\rotatebox[origin=c]{90}{\begin{minipage}{1cm}\centering MinC\\2.6\end{minipage}}}
 & Raw score & 23.6 & 31.2 & 58.7 & 61.5 & 58.0 \\
 & \Smoothing & 53.0 & \textbf{39.7} & \textbf{66.2} & 83.8 & 83.0 \\
 & \cellcolor{lightcyan}DISCODE & \textbf{\cellcolor{lightcyan}53.5} & \textbf{\cellcolor{lightcyan}39.7} & \textbf{\cellcolor{lightcyan}66.2} & \textbf{\cellcolor{lightcyan}83.8} & \textbf{\cellcolor{lightcyan}83.3} \\[-0.2em]
\bottomrule
\end{tabular}
\caption{Performance comparison across six LVLMs.
}
\label{tab:lmms}
\end{table}}

\paragraph{Generalization Across Divergence Measures}
Since the divergence term is a critical component of DISCODE, we examine its performance using different divergence measures.
Specifically, we replace the weighted KL divergence with
the Jensen–Shannon divergence, the beta divergence, the R\'{e}nyi divergence, and the standard KL divergence.
In Table \ref{tab:divergence}, we observe that DISCODE can effectively leverage these divergence measures.
The results also justify our selection of the weighted KL divergence, for which we derived an analytical solution, as the best-performing option.

{\setlength{\tabcolsep}{2.7pt}
\tablebegin
\centering
\small
{
\setlength{\tabcolsep}{2.6pt} 
\begin{tabular}{l c c c c c c c c c c}
\toprule
\multirowforfull{Divergence}
\setlength{\tabcolsep}{1.5pt}
\disableforhalf{&}
&
\datasetlist \\
\midrule
w/o divergence &
\disableforhalf{&}
\flkcom{49.6}{49.9}{39.9}{20.6}{59.6}{64.4}
\pascal{61.7}{99.7}{97.7}{72.7}{83.0}
\mdce{80.9}
\\
R\'{e}nyi divergence &
\disableforhalf{&}
\flkcom{55.6}{55.9}{40.1}{20.7}{60.8}{65.8}
\pascal{63.1}{\textbf{99.8}}{98.0}{76.8}{84.4}
\mdce{83.0}
\\
Beta divergence &
\disableforhalf{&}
\flkcom{54.9}{55.3}{40.0}{20.7}{\textbf{61.1}}{\textbf{66.0}}
\pascal{61.3}{\textbf{99.8}}{98.0}{\textbf{77.8}}{84.2}
\mdce{81.5}
\\
Jensen–Shannon div. &
\disableforhalf{&}
\flkcom{55.3}{55.6}{40.1}{20.7}{60.8}{65.7}
\pascal{61.4}{\textbf{99.8}}{98.0}{77.5}{84.1}
\mdce{81.6}
\\
Kullback-Leibler div. &
\disableforhalf{&}
\flkcom{55.2}{55.6}{\textbf{40.2}}{\textbf{20.8}}{60.5}{65.4}
\pascal{\textbf{63.9}}{\textbf{99.8}}{\textbf{98.1}}{75.5}{84.3}
\mdce{83.0}
\\
\rowcolor{lightcyan}
Weighted KLD &
\disableforhalf{&}
\flkcom{\textbf{55.7}}{\textbf{56.1}}{\textbf{40.2}}{\textbf{20.8}}{\textbf{61.1}}{\textbf{66.0}}
\pascal{62.9}{\textbf{99.8}}{97.9}{77.4}{\textbf{84.5}}
\mdce{\textbf{83.6}}
\\[-0.2em]
\bottomrule
\end{tabular}
}
\ifhalfsize
\caption{
DISCODE with various divergence measures.
}
\label{tab:divergence}
\else
\caption{Comparison of divergence measures.}
\label{tab:divergence_appendix}
\fi
\tableend

}

\section{Conclusion}
We introduced DISCODE, a novel test-time adaptive decoder for LVLM-based image captioning evaluation. By incorporating a unimodal prior distribution into the ATT loss, DISCODE robustly estimates the evaluation score distribution, thereby achieving better alignment with human judgments. We also introduced the MCEval benchmark, consisting of 18,000 image-caption pairs designed to benchmark the robustness of evaluation metrics across multiple visual domains. Our experiments demonstrated the superiority of DISCODE on MCEval and four representative real-image benchmarks, achieving state-of-the-art performance.

\paragraph{Limitations and Future Work}
Since our approach leverages latent decoder features, it cannot be applied to proprietary LVLMs like GPT-4o, which do not support feature extraction functionalities.
To further enhance performance of open-source LVLMs, extending bias mitigation techniques to tasks beyond image captioning evaluation remains a promising future research direction. We believe our work significantly advances the field of LVLM-based numerical evaluation and provides a solid foundation for future developments from both technical and dataset perspectives.

\section*{Acknowledgments}
This work was supported by DENSO IT LAB Recognition, Control, and Learning Algorithm Collaborative Research Chair (Science Tokyo). This work was also supported by JSPS KAKENHI Grant Numbers 23H00490 and 25K03135.

\bibliography{aaai2026}
\clearpage
\setcounter{secnumdepth}{2} 
\renewcommand{\thesection}{\Alph{section}}
\setcounter{section}{-1}

\vspace{25pt}
\begin{strip}
\centering{
\textbf{\LARGE DISCODE: Distribution-Aware Score Decoder for\\
\vspace{7pt}
Robust Automatic Evaluation of Image Captioning}\\
\vspace{14pt}
\textbf{\Large Supplemental Material}}
\vspace{20pt}
\end{strip}
\renewcommand{\thesection}{\Alph{section}}
\setcounter{section}{0}

\begin{table*}[t]
\normalsize
\centering
\setlength{\tabcolsep}{2pt}
\resizebox{1.005\textwidth}{!}{%
\begin{tabular}{c l c cc c cc c cc c cccc c cc} 
\toprule
& \multirow{2}{*}{\vspace{-4pt}\textbf{Metric}}  & \multirow{2}{*}{\vspace{-4pt}\textbf{FF}\hspace{3pt}} & \multicolumn{2}{c}{\textbf{Flickr8k-Expert}} & & \multicolumn{2}{c}{\textbf{Flickr8k-CF}} & & \multicolumn{2}{c}{\textbf{Composite}} & & \multicolumn{5}{c}{\textbf{Pascal-50S}} 
\\
\cmidrule{4-5} \cmidrule{7-8} \cmidrule{10-11} \cmidrule{13-17}
& & & Kendall $\tau_b$ & Kendall $\tau_c$  & & Kendall $\tau_b$ & Kendall $\tau_c$ & & Kendall $\tau_b$ & Kendall $\tau_c$ & & HC & HI & HM & MM & Mean\\
\midrule
& BLEU-1~\cite{Papineni2002BLEU} & -- & 32.2 & 32.3 & & 17.9 & 9.3 & & 29.0 & 31.3 & & 64.6 & 95.2  & 91.2 & 60.7 & 77.9 \\ %
& BLEU-4~\cite{Papineni2002BLEU} & -- & 30.6 & 30.8 & & 16.9 & 8.7 & & 28.3 & 30.6 & & 60.3 & 93.1 & 85.7 & 57.0 & 74.0 \\ %
& ROUGE~\cite{Lin2004ROUGE} & -- & 32.1 & 32.3 & & 19.9 & 10.3 & & 30.0 & 32.4 & & 63.9 & 95.0  & 92.3 & 60.9 & 78.0 \\ %
& METEOR~\cite{Banerjee2005METEOR} & -- & 41.5 & 41.8 & & 22.2 & 11.5 & & 36.0 & 38.9 & & 66.0 & 97.7  & 94.0 & 66.6 & 81.1 \\ %
& CIDEr~(Vedantam et al. 2015) & -- & 43.6 & 43.9 & & 24.6 & 12.7 & & 34.9 & 37.7 & & 66.5 & 97.9 & 90.7 & 65.2 & 80.1 \\ %
& SPICE~\cite{Anderson2016SPICE} & -- & 51.7 & 44.9 & & 24.4 & 12.0 & & 38.8 & 40.3 & & 63.6 & 96.3 & 86.7 & 68.3 & 78.7\\
\midrule
& BERT-S~\cite{Zhan2020BERTScore} & $\checkmark$ & -- & 39.2 & & 22.8 & -- & & -- & 30.1 & & 65.4 & 96.2 & 93.3 & 61.4 & 79.1 \\
& LEIC~\cite{Cui2018LEIC} & & 46.6 & -- & & 29.5 & -- & & -- & -- & & -- & -- & -- & -- & -- \\
& BERT-S++~\cite{Yi2020BERTScorepp} & $\checkmark$ & -- & 46.7 & & -- & -- & & -- & 44.9 & & 65.4 & 98.1 & 96.4 & 60.3 & 80.1 \\
& TIGEr~\cite{Jiang2019TIGEr} & $\checkmark$ & -- & 49.3 & & -- & -- & & -- & 45.4 & & 56.0 & \textbf{99.8} & 92.8 & 74.2 & 80.7 \\
& ViLBERTScore~\cite{Lee2020ViLBERTScore} & $\checkmark$ & -- & 50.1 & & -- & -- & & -- & 52.4 & & 49.9 & 99.6 & 93.1 & 75.8 & 79.6 \\
& MID~\cite{Kim2022MID} & $\checkmark$ & -- & 54.9  & & 37.3 & -- & & -- & -- & & 67.0 & 99.7 & 97.4 & 76.8 & 85.2 \\
& CLAIR~\cite{Chan2023CLAIR} & $\checkmark$ & -- & 48.3 & & 38.2 & -- & & -- & 61.0 & & 52.4 & 99.5 & 89.8 & 73.0 & 78.7\\
& RefCLIP-S~\cite{Hessel2021CLIPS} & $\checkmark$ & 52.6 & 53.0 & & 36.4 & 18.8 & & 51.2 & 55.4 & & 64.9 & 99.5  & 95.5 & 73.3 & 83.3\\ %
& RefPAC-S~\cite{Sarto2023PACS} & & 55.5 & 55.9 & & 37.6 & 19.5 & & 53.0 & 57.3 & & 67.7 & 99.6  & 96.0 & 75.6 & 84.7\\ %
& RefPAC-S++ViT-B~\cite{Sarto2024PACScorepp} & & 55.3 & 55.7 & & 37.9 & 19.6 & & 54.7 & 59.1 & & 67.2 & 99.6  & 96.2 & 74.2 & 84.5\\ %
& RefPAC-S++ViT-L~\cite{Sarto2024PACScorepp} & & -- & 57.9 & & 38.8 & -- & & -- & 61.6 & & -- & --  & -- & -- & 84.7\\ %
& Polos~\cite{Wada2024PolosPolaris} & & -- & 56.4 & & 37.8 & -- & & -- & 57.6 & & 70.0 & 99.6  & 97.4 & 79.0 & 86.5\\ %
& DENEB (ViT-L)~(Matsuda et al. 2024) & & -- & 56.8 && & 38.3 && -- & 58.2 && \textbf{76.1} & 99.7 & 97.4 & 77.9 & \textbf{87.8}\\
\midrule
& RefFLEUR~(Lee et al 2024)) & $\checkmark$ & -- & 51.9 & & 38.8 & -- & & -- & \textbf{64.2} & & 68.0 & \textbf{99.8} & 98.0 & 76.1 & 85.5\\
& G-VEval-C (GPT-4o)~\cite{Tong2025GVEval} & $\checkmark$ & \textbf{60.5} & \textbf{58.7} && 38.2 & 19.9 && -- & -- && -- & -- & -- & -- & -- \\
\rowcolor{lightcyan} & \textbf{RefDISCODE-LV (Ours)} & $\checkmark$ & 56.1 & 56.5 & & 40.4 & \textbf{20.9} & & \textbf{61.0} & \textbf{66.0} & & 70.6 & \textbf{99.8} & \textbf{98.6} & \textbf{82.1} & \textbf{87.8}\\
\rowcolor{lightcyan} & \textbf{RefDISCODE-IN (Ours)} & $\checkmark$ & 57.1 & 57.5 & & \textbf{40.5} & \textbf{20.9} & & 60.5 &  64.9 & & 68.0 & 99.7 & 97.7 & 79.7 & 86.3\\
\bottomrule
\end{tabular}
}
\caption{
Effectiveness of DISCODE in reference-based evaluation scenarios.
}
\label{tab:refdisode}
\end{table*}
\halfsizefalse
\def\multirowforfull#1{\multirow{2}{*}{#1}}
\def\disableforhalf#1{#1}
\def\tablebegin{\begin{table*}[t]}
\def\tableend{\end{table*}}
\def\tabularbegin{\begin{tabular}{l c cc c cc c cc c cc c ccccc c cc}}
\def\tabularend{\end{tabular}}
\def\datasetlist{ \multicolumn{2}{c}{\textbf{Flickr8k-Expert}} & & \multicolumn{2}{c}{\textbf{Flickr8k-CF}} & & \multicolumn{2}{c}{\textbf{Composite}} & & \multicolumn{5}{c}{\textbf{Pascal-50S}}\\
\cmidrule{3-4} \cmidrule{6-7} \cmidrule{9-10} \cmidrule{12-16}
& & \scalebox{0.9}{Kendall\hspace{1pt}$\tau_b$} & \scalebox{0.9}{Kendall\hspace{1pt}$\tau_c$} & & \scalebox{0.9}{Kendall\hspace{1pt}$\tau_b$} & \scalebox{0.9}{Kendall\hspace{1pt}$\tau_c$} & & \scalebox{0.9}{Kendall\hspace{1pt}$\tau_b$} & \scalebox{0.9}{Kendall\hspace{1pt}$\tau_c$} & & \scalebox{0.9}{HC} & \scalebox{0.9}{HI} & \scalebox{0.9}{HM} & \scalebox{0.9}{MM} & \scalebox{0.9}{Mean}
}

\def\flkcom#1#2#3#4#5#6{#1 & #2 & & #3 & #4 & & #5 & #6 & &}
\def\pascal#1#2#3#4#5{#1 & #2 & #3 & #4 & #5}
\def\pascalraw#1#2#3#4#5{#1 & #2 & #3 & #4 & #5}
\def\rice#1#2{#1 & #2 & &}
\def\mdce#1{}

\def\flkcomb#1#2#3#4#5#6{\flkcom{\textbf{#1}}{\textbf{#2}}{\textbf{#3}}{\textbf{#4}}{\textbf{#5}}{\textbf{#6}}}
\def\pascalb#1#2#3#4#5{\pascal{\textbf{#1}}{\textbf{#2}}{\textbf{#3}}{\textbf{#4}}{\textbf{#5}}}

{\setlength{\tabcolsep}{2.5pt} \def\textbfforhalf#1{
\ifhalfsize 
\textbf{#1}
\else
#1
\fi
}
\def\textbfforfull#1{\ifhalfsize#1\else\textbf{#1}\fi}
\tablebegin
\small
\centering
\resizebox{\linewidth}{!}{
\tabularbegin
\toprule
\disableforhalf{}
\multirowforfull{Method}
\disableforhalf{&}
&
\datasetlist\\
\midrule
\rowcolor{lightcyan}
DISCODE &
\disableforhalf{&}
\discodellavanextseventytwox
\\
w/o
cross-entropy term $H$ &
\disableforhalf{&}
\flkcom{54.3}{54.6}{39.5}{20.4}{58.3}{63.1}
\pascal{62.5}{\textbf{99.8}}{97.2}{75.7}{83.8}
\mdce{81.8}
\\
w/o 
divergence term $D_{\alpha}$
\disableforhalf{&}
&
\flkcom{49.6}{49.9}{39.9}{20.6}{59.6}{64.4}
\pascal{61.7}{99.7}{97.7}{72.7}{83.0}
\mdce{80.9}
\\
w/o 
adaptive definition for $\alpha$ &
\disableforhalf{&}
\flkcom{55.2}{55.6}{\textbf{40.2}}{\textbf{20.8}}{60.5}{65.4}
\pascal{\textbf{63.9}}{\textbf{99.8}}{\textbf{98.1}}{75.5}{84.3}
\mdce{83.0}
\\
\bottomrule
\tabularend
}
\ifhalfsize
\caption{
Ablation study results.
}
\label{tab:ablation}
\else
\caption{Full results for ablation study.
First two experiments omit the cross-entropy and divergence terms in Eq.~\eqref{eq:att}, respectively. Third experiment utilizes $\alpha=0.5$ to omit the adaptive definition for $\alpha$.\vspace{1em}}
\label{tab:ablation_appendix}
\small
\centering
\tabularbegin
\toprule
\multirowforfull{\textbf{Variance} $\sigma^{2}$} & & \datasetlist \\
\midrule
1.0 & &
\flkcom{55.6}{55.9}{40.2}{20.8}{60.9}{65.9}
\pascal{61.8}{99.8}{97.9}{77.2}{84.2}\\
0.5 & &
\flkcom{55.6}{55.9}{40.2}{20.8}{60.9}{65.9}
\pascal{61.8}{99.8}{97.9}{77.2}{84.2}\\
\rowcolor{lightcyan} 0.1 & &
\flkcom{55.7}{56.1}{40.2}{20.8}{61.1}{66.0}
\pascal{62.9}{99.8}{97.9}{77.4}{84.5}\\
0.05 & &
\flkcom{55.7}{56.1}{40.2}{20.8}{61.1}{66.0}
\pascal{62.9}{99.7}{97.9}{77.4}{84.5}\\
0.01 & &
\flkcom{55.7}{56.1}{40.2}{20.8}{61.1}{66.0}
\pascal{63.0}{99.7}{97.9}{76.5}{84.3}\\
\bottomrule
\tabularend
\caption{Hyperparameter study for $\sigma$ to adaptively define $\alpha$.}
\label{tab:sigma}
\fi
\tableend
}

\begin{table*}[t]
\centering
\small
\setlength{\tabcolsep}{10pt}
\begin{tabular}{llccc}
\toprule
Method & LLM & Backbone (sec) & Scoring (sec) & Overhead (\%)\\
\midrule
Raw score & LLama-3-8B & $0.241 \pm 0.03$ & $0.00 \pm 0.0$ & 0.00\\
\Smoothing & LLama-3-8B & $0.241 \pm 0.03$ & 
$5.57 \times 10^{-4} \pm 1.0 \times 10^{-4}$ & 0.23\\
DISCODE w/o Analytical Solution & LLama-3-8B & $0.241 \pm 0.03$ & 
$2.28 \times 10^{-2} \pm 1.1 \times 10^{-3}$ & 9.46\\
\rowcolor{lightcyan} DISCODE & LLama-3-8B & $0.241 \pm 0.03$ & 
$1.78 \times 10^{-3} \pm 2.8 \times 10^{-4}$ & 0.73\\
\midrule
Raw score & Vicuna-13B & $0.199 \pm 0.01$ & $0.00 \pm 0.0$ & 0.00\\
\Smoothing & Vicuna-13B & $0.199 \pm 0.01$ & 
$4.83 \times 10^{-4} \pm 1.0 \times 10^{-4}$ & 0.24\\
DISCODE w/o Analytical Solution & Vicuna-13B & $0.199 \pm 0.01$ & 
$2.29 \times 10^{-2} \pm 1.0 \times 10^{-3}$ & 11.5\\
\rowcolor{lightcyan} DISCODE & Vicuna-13B & $0.199 \pm 0.01$ & 
$1.58 \times 10^{-3} \pm 1.8 \times 10^{-4}$ & 0.79\\
\midrule
Raw score & Hermes-Yi-34B & $0.283 \pm 0.04$ & $0.00 \pm 0.0$ & 0.00\\
\Smoothing & Hermes-Yi-34B & $0.283 \pm 0.04$ & 
$4.97 \times 10^{-4} \pm 3.2 \times 10^{-4}$ & 0.18\\
DISCODE w/o Analytical Solution & Hermes-Yi-34B & $0.283 \pm 0.04$ & 
$2.30 \times 10^{-2} \pm 1.2 \times 10^{-3}$ & 8.13\\
\rowcolor{lightcyan} DISCODE & Hermes-Yi-34B & $0.283 \pm 0.04$ & 
$1.87 \times 10^{-3} \pm 3.2 \times 10^{-4}$ & 0.66\\
\midrule
Raw score & Qwen-1.5-72B  & $0.962 \pm 0.04$ & $0.00 \pm 0.0$ & 0.00\\
\Smoothing & Qwen-1.5-72B  & $0.962 \pm 0.04$ & 
$4.00 \times 10^{-4} \pm 1.0 \times 10^{-4}$ & 0.04\\
DISCODE w/o Analytical Solution & Qwen-1.5-72B  & $0.962 \pm 0.04$ & 
$1.83 \times 10^{-2} \pm 1.5 \times 10^{-3}$ & 1.90\\
\rowcolor{lightcyan} DISCODE (w/ AS) & Qwen-1.5-72B  & $0.962 \pm 0.04$ & 
$1.10 \times 10^{-3} \pm 1.1 \times 10^{-4}$ & 0.11\\
\bottomrule
\end{tabular}
\caption{Computational overhead for scoring.} \label{tab:cost}
\end{table*}


\section{Proof of Analytical Solution}

The analytical solution for the loss minimization problem stated in Eq.~\eqref{eq:minimization_problem} is shown in
Eq.~\eqref{tab:analytic}. Below, we provide the proof for it.

\begin{proof}
Suppose $\bm{h}_{T}$ be a constant vector.
For the DISCODE architecture defined in Eq.~\eqref{eq:arc},
the ATT loss is expressed as a function of the learnable parameters $W$ and $\bm{b}$ as follows:
\begin{align}
\mathcal{L}_{\text{\scalebox{0.8}{ATT}}}(W, \bm{b})
=
H(\bm{z}, p_{\text{\tiny LVLM}})
+
D_{\alpha} (\bm{z} \| q),
\end{align}
where
\begin{align}
\label{eq:11}
\bm{z} = \mathrm{softmax}( W^{\top} \bm{h}_{T} + \bm{b}).
\end{align}
For the weight matrix $W$, we have
\begin{align}
\frac{\partial \mathcal{L}_{\text{\scalebox{0.8}{ATT}}}}{\partial W_{ji}}
&=
\sum_{k} \frac{\partial \mathcal{L}_{\text{\scalebox{0.8}{ATT}}}}{\partial z_{k}} \frac{\partial z_{k}}{\partial W_{ji}}
\end{align}
\vspace{-10pt}
\begin{align}
&=
\sum_{k} \frac{\partial \mathcal{L}_{\text{\scalebox{0.8}{ATT}}}}{\partial z_{k}}
z_{k} (\delta_{ik} - z_{i}) h_{T,j}\\
&=
z_{i} h_{T,j} \left(
\frac{\partial \mathcal{L}_{\text{\scalebox{0.8}{ATT}}}}{\partial z_{i}}
 -
\sum_{k} \frac{\partial \mathcal{L}_{\text{\scalebox{0.8}{ATT}}}}{\partial z_{k}}
z_{k}
\right).
\end{align}
Setting $\partial \mathcal{L}_{\text{\scalebox{0.8}{ATT}}}/\partial W_{ji} = 0$ for all $i, j$ leads to the condition:
\begin{align}
\label{eq:cond}
\frac{\partial \mathcal{L}_{\text{\scalebox{0.8}{ATT}}}}{\partial z_{i}}
=
\sum_{k} \frac{\partial \mathcal{L}_{\text{\scalebox{0.8}{ATT}}}}{\partial z_{k}}
z_{k}~(\forall i).
\end{align}
For the bias vector $\bm{b}$, we similarly have
\begin{align}
\frac{\partial \mathcal{L}_{\text{\scalebox{0.8}{ATT}}}}{\partial b_{i}}
&=
\sum_{k} \frac{\partial \mathcal{L}_{\text{\scalebox{0.8}{ATT}}}}{\partial z_{k}} \frac{\partial z_{k}}{\partial b_{i}}\\
&=
\sum_{k} \frac{\partial \mathcal{L}_{\text{\scalebox{0.8}{ATT}}}}{\partial z_{k}}
z_{k} (\delta_{ik} - z_{i})\\
&=
z_{i} \left(
\frac{\partial \mathcal{L}_{\text{\scalebox{0.8}{ATT}}}}{\partial z_{i}}
 -
\sum_{k} \frac{\partial \mathcal{L}_{\text{\scalebox{0.8}{ATT}}}}{\partial z_{k}}
z_{k}
\right).
\end{align}
Thus, setting $\partial \mathcal{L}_{\text{\scalebox{0.8}{ATT}}}/\partial b_{i} = 0$ for all $i$ yields the same condition as in Eq.~\eqref{eq:cond}.
From Eq.~\eqref{eq:cond}, $\partial \mathcal{L}_{\text{\scalebox{0.8}{ATT}}}/\partial z_{i}$ must be constant.
Hence, there exists a constant $\Lambda$ such that
\begin{align}
\frac{1}{\alpha}
\frac{\partial \mathcal{L}_{\text{\scalebox{0.8}{ATT}}}}{\partial z_{i}}
&=
\frac{1}{\alpha}
\left(
\frac{\partial H (\bm{z} \| p_{\text{\tiny LVLM}})}{\partial z_{k}}
+
\frac{\partial D_{\alpha} (\bm{z} \| q)}{\partial z_{k}}
\right)
\label{eq:19}
\\
&=
\label{eq:20}
\Lambda_{k} + \log z_{k} + Z \equiv \Lambda~(\forall k), 
\end{align}
where
\begin{align}
\Lambda_{k}
& = \frac{1}{\alpha}
\left(
\sum_{j} V^{\top}_{jk} h_{T,j} + c_{k}\right)
+ \frac{1-\alpha}{\alpha} q(k),
\end{align}
and
\begin{align}
Z = \frac{1}{\alpha}
\log \left(
\sum_{k} \exp
\left(
\sum_{j} V^{\top}_{jk} h_{T,j} + c_{k}\right)\right) + 1.
\end{align}
Here, we assumed $p_{\text{\tiny LVLM}} = \mathrm{softmax}(V^{\top} \bm{h}_{T} + \bm{c})$ and used
\begin{align}
\frac{\partial H (\bm{z}, p_{\text{\tiny LVLM}})}{\partial z_{k}}
&=
- \log p_{\text{\tiny LVLM}}(k),\\
\frac{\partial D_{\alpha} (\bm{z} \| q)}{\partial z_{k}}
&=
(1-\alpha) q(k)
+ \alpha \log z_{k} + \alpha.
\end{align}
Then, from Eq.~\eqref{eq:20}, we obtain
\begin{align}
\label{eq:25}
z_{k} = \frac{\exp \Lambda_{k}}{\sum_{k} \exp \Lambda_{k}}.
\end{align}
Therefore, from Eq.~\eqref{eq:11} and Eq.~\eqref{eq:25}, we have
\begin{align}
\sum_{j} W_{ji}&  h_{T,j} + b_{k} \nonumber\\
&= \frac{1}{\alpha}
\left(
\sum_{j} V_{jk} h_{T,j} + c_{k}\right)
+ \frac{1-\alpha}{\alpha} q(k).
\end{align}
Finally, this provides the solutions for $W$ and $\bm{b}$:
\begin{align}
W_{ji} &= \frac{1}{\alpha} V_{ji},\\
b_{i} &= \frac{1-\alpha}{\alpha} \log q(i) + \frac{1}{\alpha} c_{i}.
\end{align}
\end{proof}

\section{Additional Experiments}

\subsection{Reference-Based DISCODE}

We investigate the effectiveness of DISCODE in reference-based scenarios. Although we introduced DISCODE as a reference-free metric, it can be naturally extended to a reference-based metric. Specifically, LVLMs can incorporate reference captions into the instruction prompts via in-context learning. We refer to this reference-based extension of DISCODE as RefDISCODE.

Table~\ref{tab:refdisode} compares RefDISCODE against state-of-the-art reference-based metrics. As shown, the results remain consistent with those observed in the reference-free experiments. RefDISCODE notably outperforms existing string-based reference metrics, such as Polos.

\subsection{Ablation and Hyperparameter Studies}

\paragraph{Ablation study}
Table~\ref{tab:ablation_appendix} shows the detailed results for the ablation study in Sec. 5.3.
As shown, removing the cross-entropy term or the divergence term led to significant performance reductions. With fixed $\alpha$, which uses the standard KL divergence, still performs well on Flickr8k-CF and Pascal-50S, but reduces the performance on Flickr8k-Expert and Composite.

\paragraph{Hyperparameter study}
Table~\ref{tab:sigma} presents a hyperparameter study for the variance hyperparameter $\sigma^{2}$ in the adaptive definition of $\alpha$. As shown, our method is robust to changes in the variance hyperparameter, with values from $0.01$ to $0.1$ yielding better results than the others.

\subsection{Efficiency of Analytical Solution}

Table~\ref{tab:cost} analyzes the computational overhead required to compute evaluation scores. As shown, using the analytical solution proposed in our method significantly reduces overhead, resulting in less than 1\% computational overhead across all models.
However, the increase in computational cost is small compared to the FLEUR baseline.

\subsection{Results on FOIL}
Table~\ref{tab:foil} compares DISCODE with the baselines on the FOIL benchmark~\cite{shekhar-etal-2017-foil}, which is specifically designed to assess a model’s ability to detect hallucinated content in image captions. 
As shown, DISCODE outperforms all baselines, including reference-based ones, demonstrating strong hallucination detection capabilities and further validating its robustness as a reference-free evaluation metric.

\begin{table}[H]
\setlength{\tabcolsep}{10pt}
\small
\centering
\begin{tabular}{lcc}
\toprule
\textbf{Metric} & \textbf{1-ref} & \textbf{4-ref} \\
\midrule
RefPAC-S~\cite{Sarto2023PACS} & 93.7 & 94.9 \\
Polos~\cite{Wada2024PolosPolaris} & 93.3 & 95.4 \\
\midrule
PAC-S~\cite{Sarto2023PACS} & 89.9 & 89.9 \\
FLEUR~\cite{Lee2024FLEUR} & 96.8 & 96.8 \\ 
G-VEval~\cite{Tong2025GVEval} & 97.8 & \textbf{98.4} \\
\rowcolor{lightcyan} DISCODE & \textbf{98.2} & 98.2 \\ 
\bottomrule
\end{tabular}
\caption{Performance comparison on FOIL.}
\label{tab:foil}
\vspace{-10pt}
\end{table}

{\setlength{\tabcolsep}{3pt}

\tablebegin
\centering
\small
\begin{tabular}{ll *{14}{c}} 

\toprule
\multirowforfull{\textbf{Metric}} 
& 
\multirowforfull{\textbf{Scale}} 
& 
\datasetlist \\
\midrule
Raw score & 1 to 5 &
\flkcom{22.3}{22.5}{30.0}{15.5}{51.2}{55.4}
\pascal{25.5}{99.1}{95.1}{23.4}{60.7}
\mdce{\red{WIP}}
\\
Raw score & 0 to 9 &
\flkcom{21.3}{21.4}{29.1}{15.0}{55.6}{60.1}
\pascal{36.3}{98.9}{97.0}{32.1}{66.1}
\mdce{\red{WIP}}
\\
Raw score & A to E &
\flkcom{29.0}{29.2}{33.2}{17.2}{53.3}{57.6}
\pascal{8.9}{99.4}{91.8}{31.3}{57.8}
\mdce{\red{WIP}}
\\
Raw score & 0.0 to 1.0 &
\rawscorellavanextseventytwo
\\
\midrule
Smoothing & 1 to 5 &
\flkcom{52.1}{52.5}{39.8}{20.5}{59.9}{64.7}
\pascal{62.0}{99.7}{98.0}{74.4}{83.5}
\mdce{\red{WIP}}
\\
Smoothing & 0 to 9 &
\flkcom{54.0}{54.4}{40.1}{20.7}{60.7}{65.6}
\pascal{62.4}{\textbf{99.8}}{97.9}{76.6}{84.1}
\mdce{\red{WIP}}
\\
Smoothing & A to E &
\flkcom{53.5}{53.8}{39.8}{20.6}{60.6}{65.5}
\pascal{61.4}{99.7}{97.5}{75.5}{83.5}
\mdce{\red{WIP}}
\\
\Smoothing & 0.0 to 1.0 &
\smoothingllavanextseventytwo
\\
\midrule
DISCODE & 
1 to 5 &
\flkcom{54.0}{54.4}{40.1}{20.7}{60.2}{65.1}
\pascal{62.8}{99.7}{\textbf{98.1}}{77.2}{\textbf{84.5}}
\mdce{\textbf{83.6}}
\\
DISCODE & 
0 to 9 &
\flkcom{55.1}{55.4}{40.1}{20.7}{60.9}{65.8}
\pascal{\textbf{64.0}}{\textbf{99.8}}{97.7}{75.1}{84.2}
\mdce{83.5}
\\
DISCODE & 
A to E &
\flkcom{54.1}{54.4}{39.8}{20.6}{60.8}{65.7}
\pascal{61.8}{99.7}{97.4}{75.7}{83.7}
\mdce{83.5}
\\
\rowcolor{lightcyan}
DISCODE & 
0.0 to 1.0 &
\flkcomb{55.7}{56.1}{40.2}{20.8}{61.1}{66.0}
\pascal{62.9}{\textbf{99.8}}{97.9}{77.4}{\textbf{84.5}}
\mdce{\textbf{83.6}}
\\
\bottomrule
\end{tabular}
\caption{Comparison of rating scales.}
\label{tab:rating_appendix}
\tableend}
{\setlength{\tabcolsep}{2.8pt}

\tablebegin
\centering
\small
{ 
\begin{tabular}{lc *{14}{c}}

\toprule
Divergence & & 
\datasetlist \\ 
\midrule
w/o divergence & & 
\flkcom{49.6}{49.9}{39.9}{20.6}{59.6}{64.4} 
\pascal{61.7}{99.7}{97.7}{72.7}{83.0} 
\mdce{80.9} 
\\
R\'{e}nyi divergence & & 
\flkcom{55.6}{55.9}{40.1}{20.7}{60.8}{65.8}
\pascal{63.1}{\textbf{99.8}}{98.0}{76.8}{84.4}
\mdce{83.0}
\\
Beta divergence & & 
\flkcom{54.9}{55.3}{40.0}{20.7}{\textbf{61.1}}{\textbf{66.0}}
\pascal{61.3}{\textbf{99.8}}{98.0}{\textbf{77.8}}{84.2}
\mdce{81.5}
\\
Jensen–Shannon div. & & 
\flkcom{55.3}{55.6}{40.1}{20.7}{60.8}{65.7}
\pascal{61.4}{\textbf{99.8}}{98.0}{77.5}{84.1} 
\mdce{81.6}
\\
Kullback-Leibler div. & & 
\flkcom{55.2}{55.6}{\textbf{40.2}}{\textbf{20.8}}{60.5}{65.4}
\pascal{\textbf{63.9}}{\textbf{99.8}}{\textbf{98.1}}{75.5}{84.3}
\mdce{83.0}
\\
\rowcolor{lightcyan}
Weighted KLD & & 
\flkcom{\textbf{55.7}}{\textbf{56.1}}{\textbf{40.2}}{\textbf{20.8}}{\textbf{61.1}}{\textbf{66.0}}
\pascal{62.9}{\textbf{99.8}}{97.9}{77.4}{\textbf{84.5}}
\mdce{\textbf{83.6}}
\\
\bottomrule
\end{tabular}
} 
\caption{Comparison of divergence measures.}
\label{tab:divergence_appendix}
\tableend}

\begin{table*}[t]
\small
\centering
\setlength{\tabcolsep}{6pt}
\begin{tabular}{lccccccccccccccc}
\toprule
\textbf{Metric} & \textbf{LLM}
& & \textbf{Real}
& & \textbf{Painting}
& & \textbf{Sketch}
& & \textbf{Quickdraw}
& & \textbf{Clipart}
& & \textbf{Infograph}
& & \textbf{Mean}\\
\midrule
Raw score & LLama-3-8B && 62.2 && 61.5 && 57.1 && 38.0 && 56.0 && 37.8 && 52.1\\
\Smoothing & LLama-3-8B && 81.9 && 83.7 && 78.1 && 65.7 && 77.0 && 74.6 && 76.8\\
\rowcolor{lightcyan} DISCODE  & LLama-3-8B && \textbf{82.5} && \textbf{84.2} && \textbf{78.4} && \textbf{66.4} && \textbf{77.8} && \textbf{75.2} && \textbf{77.4}\\[-0.2em]
\midrule
Raw score & Vicuna-13B && 61.2 && 61.9 && 55.9 && 55.7 && 55.3 && 39.7 && 55.0\\
\Smoothing & Vicuna-13B && 84.2 && 84.3 && 81.0 && 76.6 && 76.4 && 81.1 && 80.6\\
\rowcolor{lightcyan} DISCODE  & Vicuna-13B && \textbf{84.8} && \textbf{84.6} && \textbf{82.3} && \textbf{77.1} && \textbf{78.3} && \textbf{81.2} && \textbf{81.4}\\[-0.2em]
\midrule
Raw score & Hermes-Yi-34B && 71.0 && 66.9 && 61.3 && 41.6 && 62.2 && 63.1 && 61.0\\
\Smoothing & Hermes-Yi-34B && 84.2 && 83.4 && 77.9 && 75.8 && 78.0 && 79.2 && 79.7\\
\rowcolor{lightcyan} DISCODE  & Hermes-Yi-34B && \textbf{85.8} && \textbf{85.1} && \textbf{83.2} && \textbf{77.9} && \textbf{82.0} && \textbf{80.8} && \textbf{82.5}\\[-0.2em]
\midrule
Raw score & Qwen-1.5-72B && 67.9 && 62.0 && 61.7 && 46.4 && 62.0 && 69.7 && 61.6\\
\Smoothing & Qwen-1.5-72B && 86.9 && 84.3 && 83.1 && 76.3 && 82.0 && 82.3 && 82.5\\
\rowcolor{lightcyan} DISCODE  & Qwen-1.5-72B && \textbf{87.8} && \textbf{85.2} && \textbf{83.9} && \textbf{78.5} && \textbf{83.5} && \textbf{82.8} && \textbf{83.6}\\[-0.2em]
\bottomrule
\end{tabular}
\caption{Performance comparison on MCEval across four LLMs within the LLaVA-NeXT framework.\vspace{1em}}
\label{tab:mdce_llm_full}
\vspace{1em}
\setlength{\tabcolsep}{3.6pt}
\small
\centering
\begin{tabular}{lc *{14}{c}}
\toprule
\multirow{2}{*}{\textbf{Metric}} & \multirow{2}{*}{\textbf{LLM}} & \datasetlist \\ \midrule
Raw score & LLama-3-8B &
\flkcom{21.2}{21.3}{23.6}{12.2}{44.0}{47.5}
\pascalraw{33.6}{93.8}{88.6}{13.0}{57.2}
\mdce{52.1} 
\\
\Smoothing & LLama-3-8B & 
\flkcom{49.3}{49.6}{33.5}{17.3}{46.6}{50.4} 
\pascal{54.5}{98.8}{88.8}{\textbf{74.7}}{79.2} 
\mdce{76.8} 
\\
\rowcolor{lightcyan}
DISCODE & LLama-3-8B & 
\flkcomb{50.7}{51.1}{36.2}{18.7}{56.6}{61.2} 
\pascal{\textbf{59.7}}{\textbf{99.6}}{\textbf{94.7}}{73.6}{\textbf{81.9}} 
\mdce{\textbf{77.4}} 
\\
\midrule
Raw score & Vicuna-13B & 
\flkcom{24.4}{24.5}{29.8}{15.4}{55.2}{57.0} 
\pascalraw{34.2}{98.2}{91.2}{34.0}{64.4} 
\mdce{55.0} 
\\
\Smoothing & Vicuna-13B & 
\flkcom{51.8}{51.1}{38.0}{19.7}{57.3}{61.9}
\pascal{62.9}{\textbf{99.7}}{\textbf{96.0}}{75.7}{83.5} 
\mdce{80.6} 
\\
\rowcolor{lightcyan}
DISCODE & Vicuna-13B & 
\flkcomb{52.2}{52.6}{38.4}{19.9}{58.2}{62.9} 
\pascal{\textbf{64.1}}{\textbf{99.7}}{95.6}{\textbf{76.1}}{\textbf{83.9}} 
\mdce{\textbf{81.4}} 
\\
\midrule
Raw score & Hermes-Yi-34B & 
\flkcom{16.2}{16.3}{22.9}{11.8}{52.6}{54.6} 
\pascalraw{34.6}{94.4}{90.2}{22.9}{60.5} 
\mdce{61.0} 
\\
\Smoothing & Hermes-Yi-34B & 
\flkcom{54.0}{54.4}{\textbf{39.5}}{\textbf{20.4}}{\textbf{61.1}}{\textbf{66.1}} 
\pascalraw{60.9}{\textbf{99.8}}{\textbf{96.5}}{76.5}{83.4} 
\mdce{79.7} 
\\
\rowcolor{lightcyan}
DISCODE & Hermes-Yi-34B & 
\flkcomb{54.7}{55.0}{39.5}{20.4}{61.1}{66.1} 
\pascalb{63.2}{99.8}{96.5}{76.9}{84.1} 
\mdce{\textbf{82.5}} 
\\
\midrule
Raw score & Qwen-1.5-72B & 
\rawscorellavanextseventytwo 
\\
\Smoothing & Qwen-1.5-72B & 
\smoothingllavanextseventytwo 
\\
\rowcolor{lightcyan}
DISCODE & Qwen-1.5-72B & 
\discodellavanextseventytwo 
\\
\bottomrule
\end{tabular}
\caption{Performance comparison on real-image benchmarks across four LLMs within the LLaVA-NeXT framework.}
\label{tab:llms_appendix}
\end{table*}


\begin{table*}[t]
\small
\centering
\setlength{\tabcolsep}{4.5pt}
\begin{tabular}{lccccccccccccccc}
\toprule
\textbf{Metric} & \textbf{LVLM}
& & \textbf{Real}
& & \textbf{Painting}
& & \textbf{Sketch}
& & \textbf{Quickdraw}
& & \textbf{Clipart}
& & \textbf{Infograph}
& & \textbf{Mean}\\
\midrule
Raw score & InternVL-2.5-8B && 64.5 && 57.0 && 53.6 && 52.0 && 53.2 && 67.5 && 58.0\\
\Smoothing & InternVL-2.5-8B && 66.7 && 63.0 && 62.8 && 63.4 && 63.0 && 68.5 && 64.5\\
\rowcolor{lightcyan} DISCODE  & InternVL-2.5-8B && \textbf{68.5} && \textbf{64.5} && \textbf{65.8} && \textbf{64.3} && \textbf{65.6} && \textbf{70.6} && \textbf{66.5}\\
\midrule
Raw score & InternVL-2.5-78B && 77.2 && 65.7 && 62.1 && 54.2 && 65.7 && 72.9 && 66.3\\
\Smoothing & InternVL-2.5-78B && 81.1 && 75.0 && 70.5 && 66.2 && 70.7 && 80.7 && 74.0\\
\rowcolor{lightcyan} DISCODE  & InternVL-2.5-78B && \textbf{85.7} && \textbf{78.5} && \textbf{75.0} && \textbf{70.0} && \textbf{78.1} && \textbf{81.7} && \textbf{78.1}\\
\midrule
Raw score & Qwen2-VL-7B-I && \textbf{81.1} && 71.1 && 72.6 && 64.9 && 75.9 && \textbf{72.2} && 73.0\\
\Smoothing & Qwen2-VL-7B-I && \textbf{81.1} && 72.7 && \textbf{79.4} && \textbf{71.5} && 79.9 && 66.6 && \textbf{75.2}\\
\rowcolor{lightcyan} DISCODE  & Qwen2-VL-7B-I && \textbf{81.1} && \textbf{73.0} && 79.1 && 71.2 && \textbf{80.3} && 66.5 && \textbf{75.2}\\
\midrule
Raw score & Qwen2-VL-72B-I && 79.3 && 68.3 && 63.9 && 43.9 && 65.9 && 73.7 && 65.8\\
\Smoothing & Qwen2-VL-72B-I && 88.5 && \textbf{84.6} && 81.6 && 80.4 && 77.9 && 81.4 && 82.4\\
\rowcolor{lightcyan} DISCODE  & Qwen2-VL-72B-I && \textbf{89.8} && 84.5 && \textbf{83.0} && \textbf{82.2} && \textbf{81.7} && \textbf{81.6} && \textbf{83.8}\\
\midrule
Raw score & CogVLM2-C-19B && 65.4 && 59.1 &&  51.2 && 39.1 && 54.6 && 57.7 && 54.5\\
\Smoothing & CogVLM2-C-19B && 74.9 && 70.9 && 68.1 && 46.7 && 67.8 && 68.6 && 66.1\\
\rowcolor{lightcyan} DISCODE  & CogVLM2-C-19B && \textbf{77.7} && \textbf{72.6} && \textbf{69.5} && \textbf{49.0} && \textbf{72.2} && \textbf{70.8} && \textbf{68.6}\\
\midrule
Raw score & MiniCPM-V-2.6 && 69.8 && 58.6 && 44.2 && 46.0 && 59.3 && 62.7 && 58.0\\
\Smoothing & MiniCPM-V-2.6 && 88.0 && \textbf{86.6} && \textbf{83.3} && \textbf{72.4} && 79.4 && 83.0 && 83.0\\
\rowcolor{lightcyan} DISCODE  & MiniCPM-V-2.6 && \textbf{88.8} && 86.3 && 83.2 && 72.0 && \textbf{79.5} && \textbf{83.7} && \textbf{83.3}\\
\bottomrule
\end{tabular}
\caption{Performance comparison on MCEval across six LVLMs.}
\label{tab:mdce_lmm_full}
\end{table*}

{\setlength{\tabcolsep}{2.3pt}

\tablebegin
\small
\centering
\begin{tabular}{lc *{14}{c}}
\toprule
\multirow{2}{*}{\textbf{Metric}} & \multirow{2}{*}{\textbf{LVLM}} & \datasetlist \\
\midrule
Raw score & InternVL-2.5-8B &
\flkcom{26.7}{26.9}{30.4}{15.7}{50.9}{55.0}
\pascal{38.2}{97.0}{88.1}{40.7}{66.0}
\mdce{57.9}
\\
\Smoothing & InternVL-2.5-8B & 
\flkcom{56.0}{56.3}{36.7}{18.9}{54.0}{58.4}
\pascal{38.8}{98.9}{89.7}{\textbf{76.4}}{76.0}
\mdce{64.5}
\\
\rowcolor{lightcyan}
DISCODE & InternVL-2.5-8B & 
\flkcomb{57.0}{57.4}{39.2}{20.3}{55.2}{59.6}
\pascal{\textbf{54.0}}{\textbf{99.4}}{\textbf{91.4}}{76.0}{\textbf{80.2}}
\mdce{\textbf{66.5}}
\\
\midrule
Raw score & InternVL-2.5-78B &
\flkcom{41.7}{42.0}{35.7}{18.5}{55.1}{59.6}
\pascal{43.6}{98.8}{93.9}{55.1}{72.8}
\mdce{66.3}
\\
\Smoothing & InternVL-2.5-78B & 
\flkcom{56.5}{56.9}{36.4}{18.8}{59.8}{64.2}
\pascal{50.0}{99.4}{96.5}{\textbf{77.3}}{80.8}
\mdce{74.0}
\\
\rowcolor{lightcyan}
DISCODE & InternVL-2.5-78B & 
\flkcomb{57.7}{58.1}{40.1}{20.8}{60.5}{64.9}
\pascal{\textbf{60.3}}{\textbf{99.6}}{\textbf{96.7}}{77.2}{\textbf{83.5}}
\mdce{\textbf{78.1}}
\\
\midrule
Raw score & Qwen2-VL-7B-I &
\flkcom{14.4}{14.5}{39.1}{20.2}{49.7}{51.7}
\pascal{30.2}{96.6}{92.8}{20.7}{60.1}
\mdce{53.3}
\\
\Smoothing & Qwen2-VL-7B-I & 
\flkcom{52.3}{52.6}{39.5}{20.4}{49.2}{52.7}
\pascal{55.2}{99.6}{96.2}{75.0}{81.5}
\mdce{\textbf{75.2}}
\\
\rowcolor{lightcyan}
DISCODE & Qwen2-VL-7B-I & 
\flkcomb{52.5}{52.9}{39.6}{20.4}{61.3}{66.2}
\pascalb{61.9}{99.6}{96.3}{75.5}{83.3}
\mdce{\textbf{75.2}}
\\
\midrule
Raw score & Qwen2-VL-72B-I &
\flkcom{12.7}{12.8}{32.2}{16.6}{58.3}{63.0}
\pascal{33.9}{99.4}{94.0}{40.4}{66.9}
\mdce{65.8}
\\
\Smoothing & Qwen2-VL-72B-I & 
\flkcom{54.1}{54.1}{\textbf{40.0}}{\textbf{20.7}}{61.4}{66.3}
\pascal{61.0}{\textbf{99.8}}{97.4}{\textbf{76.7}}{83.7}
\mdce{82.4}
\\
\rowcolor{lightcyan}
DISCODE & Qwen2-VL-72B-I & 
\flkcomb{54.2}{54.4}{40.0}{20.7}{61.7}{66.7}
\pascalb{62.5}{99.8}{97.5}{76.7}{84.1}
\mdce{\textbf{83.8}}
\\
\midrule
Raw score & CogVLM2-C-19B &
\flkcom{13.6}{13.7}{14.3}{7.4}{28.3}{30.6}
\pascal{53.4}{82.3}{79.8}{53.7}{67.3}
\mdce{54.5}
\\
\Smoothing & CogVLM2-C-19B & 
\flkcom{38.9}{39.1}{29.8}{15.4}{41.0}{44.2}
\pascal{50.6}{\textbf{98.5}}{\textbf{96.1}}{\textbf{73.3}}{79.6}
\mdce{66.1}
\\
\rowcolor{lightcyan}
DISCODE & CogVLM2-C-19B & 
\flkcomb{40.1}{40.3}{31.9}{16.5}{49.1}{53.0}
\pascal{\textbf{53.9}}{98.4}{95.1}{73.2}{\textbf{80.2}}
\mdce{\textbf{68.6}}
\\
\midrule
Raw score & MiniCPM-V-2.6 &
\flkcom{23.5}{23.6}{31.2}{16.1}{54.3}{58.7}
\pascal{18.4}{98.6}{92.1}{36.9}{61.5}
\mdce{58.0}
\\
\Smoothing & MiniCPM-V-2.6 & 
\flkcom{52.7}{53.0}{\textbf{39.7}}{\textbf{20.5}}{\textbf{61.2}}{\textbf{66.2}}
\pascal{61.1}{\textbf{99.8}}{\textbf{97.8}}{\textbf{76.5}}{\textbf{83.8}}
\mdce{83.0}
\\
\rowcolor{lightcyan}
DISCODE & MiniCPM-V-2.6 & 
\flkcomb{53.1}{53.5}{39.7}{20.5}{61.2}{66.2}
\pascal{\textbf{61.4}}{\textbf{99.8}}{\textbf{97.8}}{76.0}{\textbf{83.8}}
\mdce{\textbf{83.3}}
\\
\bottomrule
\end{tabular}
\caption{Performance comparison across six LVLMs: InternVL-2.5-8B, InternVL-2.5-78B, Qwen2-VL-7B-I, Qwen2-VL-72B-I, CogVLM2-C-19B, and MiniCPM-V-2.6}
\label{tab:lmms_appendix}
\tableend

}

\subsection{Full Results}

\paragraph{Rating Scales}
Table~\ref{tab:rating_appendix} shows the full results corresponding to Table~\ref{tab:rating}.
These experiments were conducted by replacing 0.0 and 1.0 in the instruction prompt of Figure~\ref{fig:prompt} with the minimum and maximum scores of each rating method.
For example, when using a ten-point discrete scale, 0.0 and 1.0 were replaced with 0 and 9, respectively.
As shown in the table, DISCODE using a scale of 0.0 to 1.0 achieved the best performance in most cases.

\paragraph{Divergence Term}
Table~\ref{tab:divergence_appendix} 
shows detailed results using various divergence measures. As shown, weighted KL divergence using our dynamic weighting is the best choice.

\paragraph{Generalization Across LLMs}
Tables~\ref{tab:mdce_llm_full} and \ref{tab:llms_appendix} show the full results corresponding to Table~\ref{tab:llms}, which compares performance across four LLMs: LLama-3-8B~\cite{Dubey2024Llama3}, Vicuna-13B~\cite{vicuna2023}, Nous-Hermes-2-Yi-34B~\cite{NousResearch2024HermesYi}, and Qwen-1.5-72B~\cite{QwenTeam2024qwen1.5}. As shown, performance improvements are consistent across all domains in MCEval, as well as both Kendall's $\tau_{b}$ and $\tau_{c}$ in Flickr8k-Expert,  Flickr8k-CF and Composite.

\paragraph{Generalization Across LVLMs}
Tables~\ref{tab:mdce_lmm_full} and \ref{tab:lmms_appendix} shows the full results corresponding to Table~\ref{tab:lmms} with six state-of-the-art LLMs:
InternVL-2.5-8B,
InternVL-2.5-78B~\cite{Chen2024InternVL, Chen2024InternVL2_5},
Qwen2-VL-7B-Instruct,
Qwen2-VL-72B-Instruct~\cite{Bai2023QwenVL, Wang2024Qwen2VL},
CogVLM2-Chat-19B~\cite{Wang2023CogVLM,Hong2024CogVLM2},
Mini-CPM-V-2.6~\cite{Hu2024MiniCPM, Yao2024MiniCPMv}.
Consistent with the results for LLMs, we observed performance improvements across both Kendall's $\tau_{b}$ and $\tau_{c}$ and most subcategories in Pascal-50S.

\subsection{Qualitative Examples} 
\begin{figure*}
\centering    
\includegraphics[width=\linewidth]{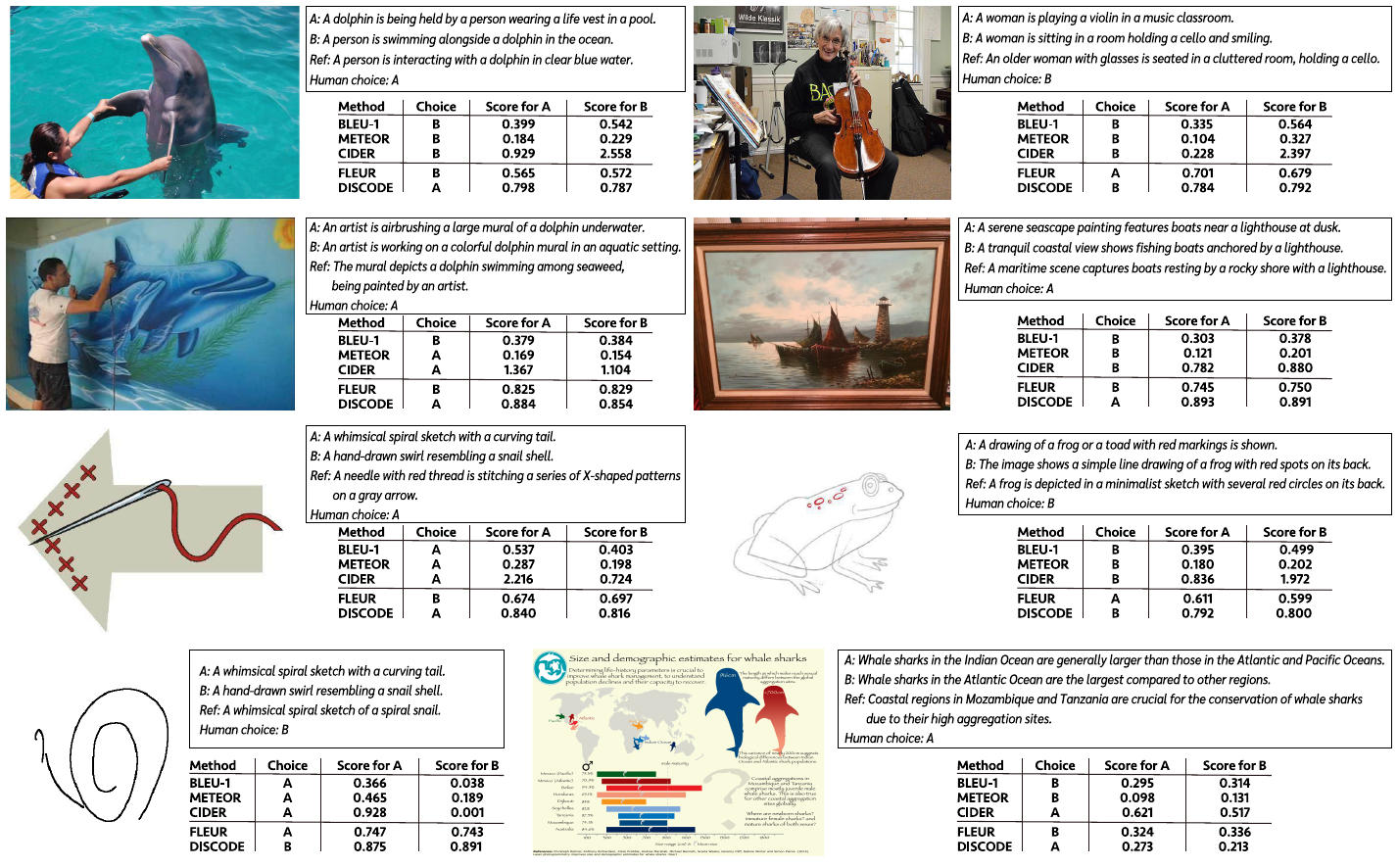}
\caption{Qualitative examples.}
\label{fig:qualitative}
\vspace{-100pt}
\end{figure*}

Figure~\ref{fig:qualitative} shows qualitative examples with evaluation scores on MCEval. As shown, DISCODE selects more accurate captions across various visual domains compared to FLEUR. It effectively identifies the key action in the image and is less influenced by commonly associated but misleading phrases. For example, in the upper-left images, DISCODE correctly assigns a higher score to the caption containing the key phrase ``being held'' over ``swimming along.'' Similarly, in the upper right image, it selects ``sitting in a room holding a cello'' instead of the commonly associated phrase ``playing a violin,'' demonstrating better contextual understanding.

DISCODE also correctly identifies key characteristics of the objects presented in the image and assigns higher scores to captions that are more specific. For instance, in the middle-right sketch image of the frog, the caption containing the more precise description ``red spots on its back'' receives a higher score than the one with ``red markings.'' Similarly, in the bottom-left infographic image about sharks, despite the presence of abundant information, various details, and distracting visuals unrelated to the infographic’s main purpose, DISCODE selects captions that accurately reflect the core content. 
Some ambiguous images such as those from the quickdraw domain remain challenging to evaluate. For instance, in the bottom-left image of an arrow, DISCODE assigns a high score to one of the candidate captions, even though both are completely unrelated to the image. Evaluating image captions for ambiguous visuals remains an open challenge for future work.

\section{Conventional Benchmarks}

Below are the details of the conventional datasets used in our experiments.

\paragraph{Flickr8k-Expert}
This dataset comprises 5,664 captions corresponding to 1,000 images sourced from Flickr. Each caption was independently evaluated by three expert annotators, who rated them on a scale from 1 to 4.

\paragraph{Flickr8k-CF}
This dataset consists of 47,830 image-caption pairs associated with 1,000 unique images. 
Each caption was evaluated by three different annotators who determined whether the caption accurately described the corresponding image, resulting in a binary “yes” or “no” judgment.
The final score for each image-caption pair is the proportion of “yes” responses, providing a measure of caption accuracy based on crowd-sourced evaluations.

\paragraph{Composite}
This dataset includes 3995 images, combining 997 images from Flickr8k, 991 images from Flickr30k~\cite{young-etal-2014-image}, and 2,007 images from MSCOCO~\cite{lin2015microsoftcococommonobjects}.
Each image is paired with three captions, and human evaluators rated each caption’s relevance to the corresponding image on a scale from 1 to 5.

\paragraph{PASCAL-50S}
This dataset comprises 1,000 images from the UIUC PASCAL Sentence Dataset, each accompanied by approximately 50 human-generated captions. It includes 4,000 caption pairs evaluated by annotators to determine which caption better describes the image.

\section{Human Annotation and User Study}
\paragraph{Annotators} Annotations for MCEval were collected via the MTurk platform. For MTurk, we recruited $81$ crowdworkers located in the United States or the United Kingdom who met strict quality criteria: at least $\geq$99\% approval rate, over $5,000$ approved HITs, and possession of the MTurk Masters qualification. The total annotation cost was approximately USD $6,500$, corresponding to an average wage of around USD $23$ per hour, ensuring fair compensation and high-quality annotations.

\paragraph{User Study} 
To verify the quality of annotations, we conducted a user study on a subset of 1,000 images via the UpWork platform.
The average human performance on MCEval on this subset was 94.2\%, indicating high alignment and annotation quality.
\end{document}